\documentclass[acmsmall]{acmart}
\usepackage{CJKutf8}
\AtBeginDocument{%
  \providecommand\BibTeX{{%
    \normalfont B\kern-0.5em{\scshape i\kern-0.25em b}\kern-0.8em\TeX}}}

\setcopyright{none}
\copyrightyear{}
\acmYear{2022}
\acmDOI{}

\usepackage{multirow}
\usepackage{lscape}
\usepackage[figuresright]{rotating}
\usepackage{float}
\usepackage{tabularx}

\usepackage[normalem]{ulem}
\useunder{\uline}{\ul}{}

\newcommand\yeon[1]{\textcolor{black}{#1}}

\newcommand\bang[1]{\textcolor{black}{#1}}
\newcommand\ziwei[1]{\textcolor{black}{#1}}
\newcommand\rita[1]{\textcolor{black}{#1}}
\newcommand\yan[1]{\textcolor{black}{#1}}

\acmJournal{CSUR}
\acmMonth{2}

\begin{document}
\settopmatter{printacmref=false}

\title{Survey of Hallucination in Natural Language Generation}

\author{Ziwei Ji}
\email{zjiad@connect.ust.hk}

\author{Nayeon Lee}
\email{nyleeaa@connect.ust.hk}
\author{Rita Frieske}
\email{rita.frieske@ust.hk}
\author{Tiezheng Yu}
\email{tyuah@connect.ust.hk}
\author{Dan Su}
\email{dsu@connect.ust.hk}
\author{Yan Xu}
\email{yxucb@connect.ust.hk}
\author{Etsuko Ishii}
\email{eishii@connect.ust.hk}
\author{Yejin Bang}
\email{yjbang@connect.ust.hk}
\author{Delong Chen}
\email{delong.chen@connect.ust.hk}
\author{Wenliang Dai}
\email{wdaiai@connect.ust.hk}
\author{Ho Shu Chan}
\email{hschanav@connect.ust.hk}
\author{Andrea Madotto}
\email{amadotto@connect.ust.hk}
\author{Pascale Fung}
\email{pascale@ece.ust.hk}

\affiliation{%
  \institution{Center for Artificial Intelligence Research (CAiRE), Hong Kong University of Science and Technology}
  \city{Clear Water Bay}
  \country{Hong Kong}
}



\renewcommand{\shortauthors}{Ziwei Ji, et al.}

\begin{abstract}
Natural Language Generation (NLG) has improved exponentially in recent years thanks to the development of sequence-to-sequence deep learning technologies such as Transformer-based language models. This advancement has led to more fluent and coherent NLG, leading to improved development in downstream tasks such as abstractive summarization, dialogue generation and data-to-text generation. However, it is also apparent that deep learning based generation is prone to hallucinate unintended text, which degrades the system performance and fails to meet user expectations in many real-world scenarios. 
To address this issue, many studies have been presented in measuring and mitigating hallucinated texts, but these have never been reviewed in a comprehensive manner before.

In this survey, we thus provide a broad overview of the research progress and challenges in the hallucination problem in NLG. The survey is organized into two parts: (1) a general overview of metrics, mitigation methods, and future directions; (2) an overview of task-specific research progress on hallucinations in the following downstream tasks, namely abstractive summarization, dialogue generation, generative question answering, data-to-text generation, machine translation, and visual-language generation; and (3) hallucinations in large language models (LLMs)~\footnote{This section was updated in Jan 2024.}.
This survey serves to facilitate collaborative efforts among researchers in tackling the challenge of hallucinated texts in NLG. 
\end{abstract}


\maketitle

\tableofcontents
\newpage

\section{Introduction}
\label{section:introduction}
Natural Language Generation (NLG)  is one of the crucial yet challenging sub-fields of Natural Language Processing (NLP). NLG techniques are used in many downstream tasks such as summarization, dialogue generation, generative question answering (GQA), data-to-text generation, and machine translation. Recently, the rapid development of NLG has captured the imagination of many thanks to the advances in deep learning technologies, especially  Transformer~\cite{vaswani2017attention}-based models like BERT~\cite{devlin2018bert}, BART~\cite{lewis2020bart}, GPT-2~\cite{radford2019language}, and GPT-3~\cite{brown2020language}. The conspicuous development of NLG tasks attracted the attention of many researchers, leading to an increased effort in the field. 

Alongside the advancement of NLG models, attention towards their limitations and potential risks has also increased.
Some early works~\cite{welleck2019neural,holtzman2019curious} focus on the potential pitfalls of utilizing the standard likelihood maximization-based objective in training and decoding of NLG models. They discovered that such likelihood maximization approaches could result in \textit{degeneration}, which refers generated output that is bland, incoherent, or gets stuck in repetitive loops. 
Concurrently, it is discovered that NLG models often generate text that is nonsensical, or unfaithful to the provided source input~\cite{vinyals2015neural, koehn2017six, rohrbach2018object, Raunak2021}. Researchers started referring to such undesirable generation as \textit{hallucination}~\cite{maynez2020faithfulness}~\footnote{The term ``hallucination'' first appeared in Computer Vision (CV) in~\citet{baker2000hallucinating} and carried more positive meanings, such as superresolution~\citep{liu2007face,baker2000hallucinating}, image inpainting~\citep{fawzi2016image}, and image synthesizing~\citep{zhang2019few-shot}. Such hallucination is something we take advantage of rather than avoid in CV. 
Nevertheless, recent works have started to refer to a specific type of error as "hallucination" in image captioning~\citep{biten2022let,rohrbach2018object} and object detection~\citep{kayhan2021hallucination,bai2009active}, which denotes non-existing objects detected or localized incorrectly at their expected position. The latter conception is similar to ``hallucination'' in NLG.}.


Hallucination in NLG is concerning because it hinders performance and raises safety concerns for real-world applications.
For instance, in medical applications, a hallucinatory summary generated from a patient information form could pose a risk to the patient. 
It may provoke a life-threatening incident for a patient if the instructions of a medicine generated by machine translation are hallucinatory.
Hallucination can also lead to potential privacy violations. \citet{carlini2020extracting} demonstrate that language models can be prompted to recover and generate sensitive personal information from the training corpus (e.g., email address, phone/fax number, and physical address). Such memorization and recovery of the training corpus is considered a form of hallucination because the model is generating text that is not ``faithful'' to the source input content (i.e., such private information does not exist in the source input).

Currently, there are many active efforts to address hallucination for various NLG tasks. Analyzing hallucinatory content in different NLG tasks and investigating their relationship would strengthen our understanding of this phenomenon and encourage the unification of efforts from different NLG fields.
However, to date, little has been done to understand hallucinations from a broader perspective that encompasses all major NLG tasks. To the best of our knowledge, existing surveys have only focused specific tasks like abstractive summarization~\cite{maynez2020faithfulness,huang2021factual} and translation~\cite{lee2018hallucinations}.  
Thus, in this paper, we present a survey of the research progress and challenges in the hallucination problem in NLG. And offer a comprehensive analysis of existing research on the phenomenon of hallucination in different NLG tasks, namely abstractive summarization, dialogue generation, generative question answering, data-to-text generation, and machine translation.
We mainly discussed hallucination of the unimodal NLG tasks that have textual input sources upon which the generated text can be assessed. We also briefly summarize hallucinations in multi-modal settings such as visual-language tasks~\citep{biten2022let,alayrac2022flamingo}.
This survey can provide researchers with a high-level insight derived from the similarities and differences of different approaches. Furthermore, given the various stages of development in studying hallucination from different tasks, the survey can assist researchers in drawing inspiration on concepts, metrics, and mitigation methods.

\paragraph{\textbf{Organization of this Survey}} 
The remainder of this survey is organized as follows. Section~\ref{section:definition} $\sim$ Section~\ref{section:futurework} provide an overview of the hallucination problem in NLG by discussing the definition and categorization, contributors, metrics, and mitigation methods of hallucinations, respectively. 
The second part of our survey discusses the hallucination problem associated with specific NLG tasks: abstractive summarization in Section~\ref{section:summarization}, dialogue generation in Section~\ref{section:dialogue}, GQA in Section~\ref{section:QA}, data-to-text generation in Section~\ref{section:data2text}, machine translation in Section~\ref{section:translation}, and VL generation in Section~\ref{section:VL}.
The third part discusses this phenomenon in LLMs in Section~\ref{section: LLM}.
Finally, we conclude the whole survey in Section~\ref{section:conclusion}.

\section{Definitions}
\label{section:definition}
In the general context outside of NLP, hallucination is a psychological term referring to a particular type of perception~\citep{fish2009perception, macpherson2013hallucination}. \citet{blom2010dictionary} define hallucination as \textbf{``a percept, experienced by a waking individual, in the absence of an appropriate stimulus from the extracorporeal world''}.
Simply put, a hallucination is an unreal perception that feels real. 
The undesired phenomenon of \textbf{``NLG models generating unfaithful or nonsensical text''} shares similar characteristics with such psychological hallucinations -- explaining the choice of terminology. Hallucinated text gives the impression of being fluent and natural despite being unfaithful and nonsensical. It appears to be grounded in the real context provided, although it is actually hard to specify or verify the existence of such contexts. Similar to psychological hallucination, which is hard to tell apart from other ``real'' perceptions, hallucinated text is also hard to capture at first glance.

Within the context of NLP, the above definition of hallucination, \textit{the generated content that is nonsensical or unfaithful to the provided source content}~\citep{filippova2020controlled,maynez2020faithfulness,parikh2020totto,zhou2021detecting}, is the most inclusive and standard.
However, there do exist variations in definition across NLG tasks, which will be further described in the later task-specific sections.


\subsection{Categorization}
\label{subsec:category}
Following the categorization from previous works~\citep{maynez2020faithfulness,huang2021factual,dziri2021neural}, there are two main types of hallucinations, namely intrinsic hallucination and extrinsic hallucination.
To explain the definition and categorization more intuitively, we give examples of each category of hallucinations for each NLG downstream task in Table \ref{Table: example}.

\begin{enumerate}
\item \textbf{Intrinsic Hallucinations}: The generated output that contradicts the source content. 
For instance, in the abstractive summarization task from Table \ref{Table: example}, the generated summary ``\textit{The first Ebola vaccine was approved in \underline{2021}}'' contradicts the source content ``\textit{The first vaccine for Ebola was approved by the FDA in \underline{2019}.}''. 

\item \textbf{Extrinsic Hallucinations}: The generated output that cannot be verified from the source content (i.e., output that can neither be supported nor contradicted by the source). 
For example, in the abstractive summarization task from Table \ref{Table: example}, the information ``\textit{China has already started clinical trials of the COVID-19 vaccine.}'' is not mentioned in source. We can neither find evidence for the generated output from the source nor assert that it is wrong.
Notably, the extrinsic hallucination is not always erroneous because it could be from factually correct external information~\cite{maynez2020faithfulness, thomson2020gold}. Such factual hallucination can be helpful because it recalls additional background knowledge to improve the informativeness of the generated text. 
However, in most of the literature, extrinsic hallucination is still treated with caution because its unverifiable aspect of this additional information increases the risk from a factual safety perspective. 
\end{enumerate}

\subsection{Task Comparison}
The previous subsection is about the definition and categorization of hallucination commonly shared by many NLG tasks. Yet, there are some task-specific differences. 

For the abstractive summarization, data-to-text, and dialogue tasks, the main difference is in what serves as the ``source'' and the level of tolerance towards hallucinations. 
 The source in abstractive summarization is the input source text that is being summarized~\cite{see2017get}, while the source in data-to-text is non-linguistic data~\cite{reiter1997building,gatt2018survey}, and the source(s) in the dialogue system is dialogue history and/or the external knowledge sentences.
Tolerance towards hallucinations is very low in both the summarization~\citep{pagnoni2021understanding} and data-to-text tasks~\citep{parikh2020totto,wang2021sketch,wang2020towards} because it is essential to provide faithful generation. In contrast, the tolerance is relatively higher in dialogue systems because the desired characteristics are not only faithfulness but also user engagement, especially in open-domain dialogue systems~\citep{huang2020challenges, ji2022vscript}. 

For the generative question answering (GQA) task, the exploration of hallucination is at its early stage, so there is no standard definition or categorization of hallucination yet. However, we can see that the GQA literature mainly focuses on ``intrinsic hallucination'' where the source is the world knowledge~\citep{li2021addressing}. Lastly, unlike the aforementioned tasks, the categorizations of hallucinations in machine translation vary within the task. Most relevant literature agrees that translated text is considered a hallucination when the source text is completely disconnected from the translated target~\cite{lee2018hallucinations,muller2020domain, Raunak2021}.
For further details, please refer to Section~\ref{section:translation}.

\subsection{Terminology Clarification}
\label{subsec:terminology}
Multiple terminologies are associated with the concept of hallucination. We provide clarification of the commonly used terminologies \textit{hallucination}, \textit{faithfulness}, and \textit{factuality} to resolve any confusion. 
\textit{Faithfulness} is defined as staying consistent and truthful to the provided source -- an antonym to "hallucination." Any work that tries to maximize faithfulness thus focuses on minimizing hallucination. For this reason, our survey includes all those works that address the faithfulness of machine-generated outputs. 
\textit{Factuality} refers to the quality of being actual or based on fact. Depending on what serves as the ``fact'', "factuality" and "faithfulness" may or may not be the same. \citet{maynez2020faithfulness} differentiate "factuality" from "faithfulness" by defining the ``fact'' to be the world knowledge.
In contrast, \citet{dong2020multi} use the source input as the ``fact'' to determine the factual correctness, making "factuality" indistinguishable from "faithfulness". 
In this paper, we adopt the definition from \citet{maynez2020faithfulness} because we believe having such a distinction between source knowledge and world knowledge provides a more clear understanding. 

Note that the judging criteria for what is considered faithful or hallucinated (i.e., the definition of hallucination) can differ across tasks. For more details of these variation definitions, you can find in the later task-specific sections.


\begin{sidewaystable}[]
\centering
\resizebox{0.92\textwidth}{!}{%
\footnotesize
\begin{tabular}{clclccccccl}
\toprule
\textbf{Task}                           & \multicolumn{1}{c}{\textbf{Sub-Task}}                                                         & \textbf{Type}              & \multicolumn{7}{c}{\textbf{Source}}                                                                                                                                                                                                                                                                                                                                                                                                                & \multicolumn{1}{c}{\textbf{Output}}                                                                                                                                                                                                                                                                                                                          \\ \hline
\multirow{2}{*}{\begin{tabular}[c]{@{}c@{}}Abstractive\\Summarization\\~\citep{pagnoni2021understanding}\end{tabular}}       &                                                                                               & Intrinsic                  & \multicolumn{7}{l}{\begin{tabularx}{0.35\textwidth}{X}The first vaccine for Ebola was approved by the FDA in 2019 in the US, five years after the initial outbreak in 2014. To produce the vaccine, scientists had to sequence the DNA\end{tabularx}}                                                                                                                                                                                                 & \begin{tabularx}{0.35\textwidth}{X}The first Ebola vaccine was approved in \textbf{2021}.\end{tabularx}                                                                                                                                                              \\
                              &                                                                                               & Extrinsic                  & \multicolumn{7}{l}{\begin{tabularx}{0.35\textwidth}{X}of Ebola, then identify possible vaccines, and finally show successful clinical trials. Scientists say a vaccine for COVID-19 is unlikely to be ready this year, although clinical trials have already started.\end{tabularx}}                                                                                                                                                               & \begin{tabularx}{0.35\textwidth}{X}\textbf{China} has already started clinical trials of the COVID-19 vaccine. \end{tabularx}                                                                                                                                                                                                                                                                                         \\ \hline
\multirow{8}{*}{Dialogue}      & \multicolumn{1}{c}{\multirow{2}{*}{\begin{tabular}[c]{@{}c@{}}Task-\\ oriented\\~\cite{li-etal-2020-slot}\end{tabular}}} & Intrinsic                  & \multicolumn{7}{l}{\multirow{2}{*}{\begin{tabularx}{0.35\textwidth}{X}inform (NAME = pickwick hotel, PRICERANGE = moderate)\end{tabularx}}}                                                                                                                                                                                                                                                                                                                                                          & \begin{tabularx}{0.35\textwidth}{X}the hotel named pickwick hotel is in a \textbf{high} price range.\\ \\\end{tabularx}                                                                                                                                                                                                                                                                                                    \\
                              & \multicolumn{1}{c}{}                                                                          & Extrinsic                  & \multicolumn{7}{l}{}                                                                                                                                                                                                                                                                                                                                                                                                                               & \begin{tabularx}{0.35\textwidth}{X}the pickwick hotel \textbf{in san diego} is a moderate price range\\\\ \end{tabularx}                                                                                                                                                                                                                                                                                               \\ \cline{2-11} 
                              & \multicolumn{1}{c}{\multirow{2}{*}{\begin{tabular}[c]{@{}c@{}}Open-\\ domain \\\cite{santhanam2021rome}\end{tabular}}}   & Intrinsic                  & \multicolumn{7}{l}{\begin{tabularx}{0.35\textwidth}{X}Dialog History: \\ Speaker 1: What do you think about Murray? \\ Speaker 2: I think Murray is a great player he just needs to stay healthy in order to compete more. Who do you like best?\end{tabularx}} &\begin{tabularx}{0.35\textwidth}{X} Speaker 1: I like \textbf{Roger Nadal} the most!\\\\\\\ \end{tabularx}                                                                                                                                                                                                                                                                                                                                  \\
                              & \multicolumn{1}{c}{}                                                                          & Extrinsic                  & \multicolumn{7}{l}{\begin{tabularx}{0.35\textwidth}{X}Knowledge: In tennis, the term Big Four refers to the quartet of men’s singles players comprising Roger Federer, Rafael Nadal, Novak Djokovic, and Andy Murray.\end{tabularx}}                                                                                                                                                                                                                                                                                                                                                                                                                            & \begin{tabularx}{0.35\textwidth}{X}Speaker 1: I like Djokovic. He has played \textbf{in the top ten singles players of the world}.\end{tabularx}                                                                                                                                                                                                                                                                                 \\ \hline
\multirow{6}{*}{\begin{tabular}[c]{@{}c@{}}Generative QA\\~\citep{nguyen2016ms}\end{tabular}} &                                                                                               & Intrinsic                  & \multicolumn{7}{l}{\begin{tabularx}{0.35\textwidth}{X}Question: dow jones industrial average please? \end{tabularx}}                                                                                                                                                                                                                                                                                                                                                                                  & \begin{tabularx}{0.35\textwidth}{X}Answer: The Dow Jones Industrial Average(DJIA) is an index of \textbf{30 major U.S.stock indexes}. The DJIAwas created in 1896 by Charles Dow, a statistician and editor of The Wall Street Journal. The index was first published in the mid-1880s, and \textbf{was last updated on May 26,1896.}\end{tabularx}                         \\
                              &                                                                                               & Extrinsic                  & \multicolumn{7}{l}{Question: definition of sadduction}                                                                                                                                                                                                                                                                                                                                                                                             & \begin{tabularx}{0.35\textwidth}{X}Answer: Sadducees. noun. \textbf{The definition of a sadducee is a person who acts in a deceitful or duplicitous manner. An example of a Sadduceee is a politician who acts deceitfully in order to gain political power}.1 a  member of a Jewish sect that was active during the Second Temple.\end{tabularx}                        \\ \hline
\multirow{3}{*}{\begin{tabular}[c]{@{}c@{}}Data2text\\~\cite{wiseman2017challenges}\end{tabular}} &                                                                                               & \multirow{2}{*}{Intrinsic} & \multicolumn{1}{c}{TEAM}                                  & CITY                                  & WIN                              & LOSS                              & PTS                              & FG\_PCT                              & BLK                             & \multirow{2}{*}{\begin{tabularx}{0.35\textwidth}{X}The Houston Rockets \textbf{(18-4)} defeated the Denver Nuggets (10-13) 108-96 on Saturday.\end{tabularx}}                                                                                                                                                                                          \\ \cline{4-10}
                                                                                                                  &                                                                                               &                            & \multicolumn{1}{c}{Rockets}                               & Houston                               & 18                               & 5                                 & 108                              & 44                                   & 7                               &                                                                                                                                                                                                                                                                                                                                         \\
                                                                                                                  &                                                                                               & Extrinsic                  & \multicolumn{1}{c}{Nuggets}                               & Denver                                & 10                               & 13                                & 96                               & 38                                   & 7                               & \begin{tabularx}{0.35\textwidth}{X} \textbf{Houston has won two straight games and six of their last seven.}\end{tabularx}                                                                                                                                                                                                                                                                         \\ \hline

\multirow{2}{*}{\begin{tabular}[c]{@{}c@{}}Translation\\~\cite{zhou2021detecting}\end{tabular}}  &                                                                                               & Intrinsic                 & \multicolumn{7}{l}{\begin{tabularx}{0.35\textwidth}{X} \begin{CJK*}{UTF8}{gbsn}迈克周四去书店。\end{CJK*} (Michael went to the bookstore on Thursday.) \end{tabularx}}                                                                                                                                                                                                                                                                               & \begin{tabularx}{0.35\textwidth}{X}\textbf{Jerry didn't go} to the bookstore.\end{tabularx}                                                                                                                                                                                                           \\
                              &                                                                                               & Extrinsic                & \multicolumn{7}{l}{\begin{tabularx}{0.35\textwidth}{X}\begin{CJK*}{UTF8}{gbsn}迈克周四去书店。\end{CJK*} (Michael went to the bookstore on Thursday.) \end{tabularx}}                                                                                                                                                                                                                                                                                                                 & \begin{tabularx}{0.35\textwidth}{X}Michael \textbf{happily} went to the bookstore \textbf{with his friend.} \end{tabularx}                                                                                                                                                                         \\ \bottomrule
\end{tabular}
}
\caption{Examples of each category of hallucinations for each task.
 In the Data2Text task, H/A: home/away, MIN: minutes, PTS: points, REB: rebounds, AST: assists, BLK: blocks, FG\_PCT: field goals percentage. The examples for VL tasks are shown in Figure~\ref{fig:captioning_examples} and Figure~\ref{fig:vqa_examples}}

\label{Table: example}
\end{sidewaystable}


		

\begin{table*}[]
\resizebox{\linewidth}{!}{

\footnotesize
\centering
\begin{tabular}{cccl}
\toprule
\multicolumn{1}{c}{}                                                          & \textbf{Category}                                                                   & \multicolumn{1}{c}{\textbf{Task}} & \multicolumn{1}{c}{\textbf{Works}} \\ \midrule
\multirow{16}{*}{\begin{tabular}[c]{@{}c@{}}\textbf{Automatic}\\ \textbf{Metrics}\end{tabular}}                                                  & \multirow{4}{*}{\begin{tabular}[c]{@{}c@{}}Statistical \end{tabular}}                                                         & Dialogue                    & \citet{shuster2021retrieval}  \\\cline{3-4}   
                                                                              &                                                                                  & Data2Text                 &  \citet{dhingra2019handling,wang2020towards}  \\     \cline{3-4} 
                                                                              &
                                                                              
                                                                       &
                                         Translation                &  \citet{Martindale2019IdentifyingFI}  \\   \cline{3-4}  
                                         & & Captioning & \citet{rohrbach2018object}\\
                                         \cline{2-4} 
                                         &
                                         \multirow{10}{*}{\begin{tabular}[c]{@{}c@{}}Model-\\ based\end{tabular}}                                                        &     \begin{tabular}[c]{@{}c@{}}Abstractive\\Summarization\end{tabular}                         & \begin{tabular}[c]{@{}l@{}}\citet{nan2021entity,durmus2020feqa,wang2020asking,kryscinski2020evaluating}\\\citet{goodrich2019assessing,pagnoni2021understanding,gabriel-etal-2021-go,zhou2021detecting}\\\citet{laban2021summac,falke2019ranking,mishra2021looking,scialom2021questeval}\end{tabular}     \\\cline{3-4} 
                                                                              &                                                                                  &Dialogue                    &     \begin{tabular}[c]{@{}l@{}}\citet{li-etal-2020-slot,balakrishnan-etal-2019-constrained,honovich2021q} \\ \citet{dziri2021evaluating,gupta2021dialfact,santhanam2021rome}\end{tabular} \\\cline{3-4} 
                                                                              &                                                                                  & Generative QA                  &  \begin{tabular}[c]{@{}l@{}}\citet{sellam2020bleurt}$^*$,\citet{zhang2020optimizing}$^*$,\citet{durmus2020feqa}$^*$\\\citet{wang2020asking}$^*$, \citet{su2022read}\end{tabular}   \\\cline{3-4} 
                                                                              &                                                                                  & Data2Text                 &    
                                                                              \begin{tabular}[c]{@{}l@{}}\citet{liu2021towards,wiseman2017challenges, dusek-kasner-2020-evaluating}\\\citet{tian2020sticking,filippova2020controlled,rebuffel2021data} \end{tabular} \\\cline{3-4}  
                                                                              &                                                                                  & Translation               &   
                                                   \begin{tabular}[c]{@{}l@{}}\citet{kong2019neural, lee2018hallucinations,tu2016modeling}\\\citet{feng2020modeling,garg2019jointly,zhou2021detecting} \\\citet{Raunak2021,parthasarathi2021want} \end{tabular}                           \\\cline{3-4} 
                                                                              &                                                                                  & Task-Agnostic               &    \citet{goyal2020evaluating,zhou2021detecting,liu2021token} \\  
\midrule
\multirow{22}{*}{\begin{tabular}[c]{@{}c@{}}\textbf{Mitigation}\\ \textbf{Method}\end{tabular}} & 
                                                                             \multirow{8}{*}{\begin{tabular}[c]{@{}c@{}}Data-\\Related\end{tabular}}        &  \begin{tabular}[c]{@{}c@{}}Abstractive\\Summarization\end{tabular}                    &  \begin{tabular}[c]{@{}l@{}}\citet{nan2021entity,cao2018faithful,zhu2021enhancing}\\\citet{gunel2020mind}\end{tabular} \\ \cline{3-4} 
                                                                              &                                                                                  & Dialogue                    & \begin{tabular}[c]{@{}l@{}}\citet{shen2021identifying,wu2021controllable,honovich2021q}\\ \citet{santhanam2021rome,shuster2021retrieval} \end{tabular}\\\cline{3-4} 
                                                                              &                                                                                  & Generative QA                        &        \citet{yin2016neural, bi2019incorporating,fan2019using}                  \\\cline{3-4} 
                                                                              &                                                                                 & Data2Text                 &  \begin{tabular}[c]{@{}l@{}}\citet{parikh2020totto,wang2019revisiting,nie2019simple,liu2021towards}\\ \citet{rebuffel2022controlling, nie2018operation}\end{tabular} \\\cline{3-4} 
                                                                              &                                                                                  & Translation               &  \begin{tabular}[c]{@{}l@{}}\citet{Raunak2021,lee2018hallucinations} \\ \citet{ junczysdowmunt2019dual,briakou2021} \end{tabular}  \\ \cline{3-4} 
                                                                              & & Captioning & \citet{biten2022let} \\  \cline{2-4}

                                                                            & \multirow{12}{*}{\begin{tabular}[c]{@{}c@{}}Modeling\\and\\ Inference\end{tabular}}     &  \begin{tabular}[c]{@{}c@{}}Abstractive\\Summarization\end{tabular}                    & \begin{tabular}[c]{@{}l@{}}\citet{huang2020knowledge,li2018ensure,song2020joint}\\\citet{cao2020factual,aralikatte2021focus,cao2021cliff} \\\citet{chen2021improving,albrecht-hwa-2007-examination,zhao2020reducing}\end{tabular}
                                                                         \\\cline{3-4} 
                                                                              &                                                                                  &Dialogue                    & 
                                                                              \begin{tabular}[c]{@{}l@{}}\citet{rashkin2021increasing, balakrishnan-etal-2019-constrained,li-etal-2020-slot}\\\citet{dziri2021neural} \end{tabular}        \\\cline{3-4} 
                                                                              &                                                                                  & Generative QA                        &          \begin{tabular}[c]{@{}l@{}}\citet{li2021addressing, krishna2021hurdles, fan2019using}\\\citet{nakano2021webgpt, su2022read} \end{tabular}                  \\\cline{3-4} 
                                                                              &                                                                                  & Data2Text                 & 
                                                                              \begin{tabular}[c]{@{}l@{}}\citet{liu2021towards,xu2021agggen,tian2020sticking, wang2021sketch,wang2020towards}\\\citet{su2021plan,rebuffel2022controlling,filippova2020controlled,xiao2021hallucination}\\\citet{puduppully2021data}\end{tabular}  \\\cline{3-4} 
                                                                              &                                                                                  & Translation               & 

                                                                              \begin{tabular}[c]{@{}l@{}}\citet{feng2020modeling,weng2020towards, lee2018hallucinations}\\\citet{Li2021_TERM,Raunak2021,wang2020exposure}\\\citet{  zhou2021detecting,bengio2015scheduled}\\\citet{Goyal2017,Xu2019} \end{tabular} \\ \cline{3-4}
                                                                              & & Captioning & \citet{xiao2021hallucination,Dai2022PlausibleMN} \\
                                                                              \bottomrule
\end{tabular}}
\caption{Evaluation metrics and mitigation methods for each task. *The hallucination metrics are not specifically proposed for generative question answering (GQA), but they can be adapted for that task. } 
\label{Table: overall}
\end{table*}

\section{Contributors to Hallucination in NLG}
\label{section:contributors}
\subsection{Hallucination from Data}
The main cause of hallucination from data is source-reference divergence. This divergence happens 1) as an artifact of heuristic data collection or 2) due to the nature of some NLG tasks that inevitably contain such divergence. When a model is trained on data with source-reference(target) divergence, the model can be encouraged to generate text that is not necessarily grounded and not faithful to the provided source.

\paragraph{Heuristic data collection}
When collecting large-scale datasets, some works heuristically select and pair real sentences or tables as the source and target~\cite{lebret2016neural, wiseman2017challenges}. As a result, the target reference may contain information that cannot be supported by the source~\cite{wang2019revisiting, parikh2020totto}. 
For instance, when constructing WIKIBIO~\cite{lebret2016neural}, a dataset for generating biographical notes based on the infoboxes of Wikipedia, the authors took the Wikipedia infobox as the source and the first sentence of the Wikipedia page as the target ground-truth reference. However, the first sentence of the Wikipedia article is not necessarily equivalent to the infobox in terms of the information they contain. Indeed, \citet{dhingra2019handling} points out that 62\% of the first sentences in WIKIBIO have additional information not stated in the corresponding infobox. Such mismatch between source and target in datasets can lead to hallucination.

Another problematic scenario is when duplicates from the dataset are not properly filtered out. It is almost impossible to check hundreds of gigabytes of text corpora manually. \citet{lee2021deduplicating} show that duplicated examples 
from the pretraining corpus bias the model to favor generating repeats of the memorized phrases from the duplicated examples.

\paragraph{Innate divergence} Some NLG tasks by nature do not always have factual knowledge alignment between the source input text and the target reference, especially those that value diversity in generated output. 
For instance, it is acceptable for open-domain dialogue systems to respond in chit-chat style, subjective style~\cite{rashkin2021increasing}, or with a relevant fact that is not necessarily present in the user input, history or provided knowledge source -- this improves the engagingness and diversity of the dialogue generation. However, researchers have discovered that such dataset characteristic leads to inevitable extrinsic hallucinations.

\subsection{Hallucination from Training and Inference}
As discussed in the previous subsection, source-reference divergence existing in dataset is one of the contributors of hallucination. However, \citet{parikh2020totto} show that hallucination problem still occurs even when there is very little divergence in dataset. This is because there is another contributor of hallucinations -- training and modeling choices of neural models~\cite{vinyals2015neural, koehn2017six, rohrbach2018object, Raunak2021}.


\paragraph{Imperfect representation learning}
The encoder has the role of comprehending and encoding input text into meaningful representations. An encoder with a defective comprehension ability could influence the degree of hallucination~\citep{parikh2020totto}.
When encoders learn wrong correlations between different parts of the training data, it could result in erroneous generation that diverges from the input~\cite{tian2020sticking, feng2020modeling,aralikatte2021focus,li2018ensure}.


\paragraph{Erroneous decoding}
The decoder takes the encoded input from the encoder and generates the final target sequence. Two aspects of decoding contribute to hallucinations. First, decoders can attend to the wrong part of the encoded input source, leading to erroneous generation~\cite{tian2020sticking}. 
Such wrong association results in generation with facts mixed up between two similar entities~\cite{shuster2021retrieval,dziri2021neural}.
Second, the design of the decoding strategy itself can contribute to hallucinations. 
\citet{dziri2021neural} illustrate that a decoding strategy that improves the generation diversity, such as top-k sampling, is positively correlated with increased hallucination. We conjecture that deliberately added ``randomness'' by sampling from the top-k samples instead of choosing the most probable token increase the unexpected nature of the generation, leading to a higher chance of containing hallucinated content.

\paragraph{Exposure Bias}
Regardless of decoding strategy choices, the exposure bias problem~\cite{bengio2015scheduled, ranzato2016sequence}, defined as the discrepancy in decoding between training and inference time, can be another contributor to hallucination. 
It is common practice to train the decoder with teacher-forced maximum likelihood estimation (MLE) training, where the decoder is encouraged to predict the next token conditioned on the ground-truth prefix sequences. However, during the inference generation, the model generates the next token conditioned on the historical sequences previously generated by itself~\citep{he2021exposure}. Such a discrepancy can lead to increasingly erroneous generation, especially when the target sequence gets longer.   

\paragraph{Parametric knowledge bias}
Pre-training of models on a large corpus is known to result in the model memorizing knowledge in its parameters~\cite{petroni2019language,roberts2020much,madotto2020language}. This so-called parametric knowledge helps improve the performance of downstream tasks but also serves as another contributor to hallucinatory generation. Large pre-trained models used for downstream NLG tasks are powerful in providing generalizability and coverage, but \citet{longpre2021entity} have discovered that such models prioritize parametric knowledge over the provided input. In other words, models that favor generating output with their parametric knowledge instead of the information from the input source can result in the hallucination of excess information in the output. 
On the other hand, current research works~\cite{yin-etal-2023-large, burns2022discovering, rajpurkar2018know, kadavath2022language, Manakul2023SelfCheckGPT} highlight a discrepancy between surface realization and inherent knowledge of the model in NLG tasks. Models can realize they are generating something hallucinated in some way. 

\section{Metrics Measuring Hallucination}
\label{section:metric}
Recently, various studies have illustrated that most conventional metrics used to measure the quality of writing are not adequate for quantifying the level of hallucination~\cite{reiter2018structured}. It has been shown that state-of-the-art abstractive summarization systems, evaluated with metrics such as ROUGE, BLEU, and METEOR, have hallucinated content in 25\% of their generated summaries~\cite{falke2019ranking}. A similar phenomenon has been shown in other NLG tasks, where it has been discovered that traditional metrics have a poor correlation with human judgment in terms of the hallucination problem~\cite{krishna2021hurdles, dhingra2019handling,durmus2020feqa,honovich2021q}. 
Therefore, there are active research efforts to define effective metrics for quantifying hallucination. 
\ziwei{FRANK~\citep{pagnoni2021understanding} surveys the faithfulness metrics for summarization and compares these metrics' correlations with human judgments. To assess the example-level accuracy of metrics in diverse tasks, TRUE~\citep{honovich2022true} reports their Area Under the ROC Curve (ROC AUC) in regard to hallucinated example detection.
}

\subsection{Statistical Metric}
\label{section:statistical_metric}
One of the simplest approaches is to leverage lexical features (n-grams) to calculate the information overlap and contradictions between the generated and the reference texts -- the higher the mismatch counts, the lower the faithfulness and thus the higher the hallucination score. 

Given that many traditional metrics leverage the target text as the ground-truth reference (e.g., ROUGE, BLEU, etc.), \citet{dhingra2019handling} build upon this idea and propose PARENT (Precision And Recall of Entailed n-grams from the Table)~\footnote{Note that PARENT is a general metric like ROUGE and BLEU, not only constrained to hallucination}, a metric which can also measure hallucinations using \textit{both} the source and target text as references.
Particularly, PARENT n-gram lexical entailment matches generated text with both the source table and target text. The F1-score that combines the precision and recall of the entailment reflects the accuracy of the table-to-text task.
The source text is additionally used because it is not guaranteed that the output target text contains the complete set of information available in the input source text.

It is common for NLG tasks to have multiple plausible outputs from the same input, which is known as one-to-many mapping~\citep{su2020diversifying,guan2020union}. 
In practice, however, covering all the possible outputs is too expensive and almost impossible. Thus, many works simplify the hallucination evaluation setup by relying on the source text as the sole reference. Their metrics just focus on the information referred by input sources to measure hallucinations, especially intrinsic hallucinations.
For instance, \citet{wang2020towards} propose PARENT-T, which simplifies PARENT by only using table content as the reference. 
Similarly, Knowledge F1~\cite{shuster2021retrieval} -- a variant of unigram F1 -- has been proposed for knowledge-grounded dialogue tasks to measure the overlap between the model’s generation and the knowledge used to ground the dialogue during dataset collection. 

Furthermore, \citet{Martindale2019IdentifyingFI} proposed a bag-of-vectors sentence similarity (BVSS) metric for measuring sentence adequacy in machine translation, that only refers to the target text. This statistical metric helps to determine whether the MT output has a different amount of information than the translation reference. 

Although simple and effective, one potential limitation of lexical matching is that it can only handle lexical information. Thus, it fails to deal with syntactic or semantic variations~\cite{sellam2020bleurt}. 

\subsection{Model-based Metric}
\label{subsec:model_based_metric}

\bang{Model-based metrics leverage neural models to measure the hallucination degree in the generated text. They are proposed to handle more complex syntactic and even semantic variations. The model-based metrics comprehend the source and generated texts and detect the knowledge/content mismatches.} However, the neural models can be subject to errors that can propagate and adversely affect the accurate quantification of hallucination.

\subsubsection{Information Extraction (IE)-based} It is not always easy to determine which part of the generated text contains the knowledge that requires verification. IE-based metrics use IE models to represent the knowledge in a simpler relational tuple format (e.g., \textit{subject, relation, object}), then verify against relation tuples extracted from the source/reference. 
Here, the IE model is identifying and extracting the ``facts" that require verification. In this way, words containing no verifiable information (e.g., stopwords, conjunctions, etc) are not included in the verification step.

For example, ground-truth reference text \texttt{``Brad Pitt was born in 1963''} and generated text \texttt{``Brad Pitt was born in 1961''} will be mapped to the relation triples \texttt{(Brad Pitt, born-in, 1963)} and \texttt{(Brad Pitt, born-in, 1961)} respectively~\footnote{This is an example from~\citep{goodrich2019assessing}}. The mismatch between the dates (1963$\neq$1961) indicates that there is hallucination. 
One limitation associated with this approach is the potential error propagation from the IE model. 

\subsubsection{QA-based} This approach implicitly measures the knowledge overlap or consistency between the generation and the source reference. This is based on the intuition that similar answers will be generated from a same question if the generation is factually consistent with the source reference. It is already put in use to evaluate hallucinations in many tasks, such as summarization \ziwei{~\cite{wang2020asking, durmus2020feqa, scialom2021questeval}, dialogue~\citep{honovich2021q}, and data2text generation~\citep{rebuffel2021data}}.

QA-based metric that measures the faithfulness of the generated text is consisted of three parts: First, given a generated text, a question generation (QG) model generates a set of question-answer pairs. 
Second, a question answering (QA) model answers the generated questions given a ground-truth source text as the reference (containing knowledge). Lastly, the hallucination score is computed based on the similarity of the corresponding answers.

Similar to the IE-based metrics, the limitation of this approach is the potential error that might arise and propagated from either the QG model or the QA model.

\subsubsection{Natural Language Inference (NLI) Metrics} There are not many labelled datasets for hallucination detection tasks, especially at the early stage when the hallucination problem starts to gain attention. As an alternative, many works leverage the NLI dataset to tackle hallucinations. Note that NLI is a task that determines whether a “hypothesis” is true (entailment), false (contradiction), or undetermined (neutral) given a “premise”. These metrics are based on the idea that only the source knowledge reference should entail the entirety of the information in faithful and hallucination-free generation ~\cite{williams2018broad, falke2019ranking,mishra2021looking,kryscinski2020evaluating,dusek-kasner-2020-evaluating,huang2021factual,honovich2021q, laban2021summac, dziri2021evaluating}. 
More specifically, NLI-based metrics define the hallucination/faithfulness score to be the entailment probability between the source and its generated text, also known as the percentage of times generated text entails, neutral to, and contradicts the source. 

According to~\citet{honovich2021q}, NLI-based approaches are more robust to lexical variability than token matching approaches such as IE-based and QA-based metrics. Nevertheless, as illustrated by~\citet{falke2019ranking}, off-the-shelf NLI models tend to transfer poorly to the abstractive summarization task. Thus, there is a line of research in improving and extending the NLI paradigm specifically for hallucination evaluation purposes~\cite{falke2019ranking,dziri2021evaluating}. Apart from generalizability, \citet{goyal2020evaluating} point out the potential limitation of using sentence-level entailment models, namely their incapability to pinpoint and locate which parts of the generation are erroneous. In response, the authors propose a new dependency-level entailment and attempt to identify factual inconsistencies in a more fine-grained manner.

\subsubsection{Faithfulness Classification Metrics}
To improve upon NLI-based metrics, task-specific datasets are constructed to improve from the NLI-based metrics.
\citet{zhou2021detecting,liu2021token} constructed syntactic data by automatically inserting hallucinations into training instances. 
\ziwei{\citet{santhanam2021rome} and \citet{honovich2021q} construct new corpora for faithfulness classification in dialogue responses. They manually annotate the Wizard-of-Wikipedia dataset~\citep{dinan2018wizard}, a knowledge grounded dialog dataset, by judging whether each response is hallucinated.}

Faithfulness specific datasets can be better than NLI datasets because entailment or neutral labels of NLI datasets and faithfulness are not equivalent. 
For example, the hypothesis ``Putin is U.S. president'' can be considered to be either neutral to or entailed from the premise ``Putin is president''. However, from the faithfulness perspective, the hypothesis contains unsupported information ``U.S.'', which is deemed to be hallucination.

\subsubsection{LM-based Metrics} 
These metrics leverage two language models (LMs) to determine if each token is supported or not:
An unconditional LM is only trained on the targets (ground-truth references) in the dataset, while a conditional language model $LM_x$ is trained on both source and target data.
It is assumed that the next token is inconsistent with the input if unconditional LM gets a smaller loss than conditional $LM_x$ during forced-path decoding~\cite{filippova2020controlled, tian2020sticking}. 
We classify the generated token as hallucinatory if the loss from LM is lower. The ratio of hallucinated tokens to the total number of target tokens $|y|$ can reflect the hallucination degree.

\subsection{Human Evaluation} 
\label{subsec:human_evaluation}

Due to the challenging and imperfect nature of the current automatic evaluation of hallucinations in NLG, human evaluation~\citep{shuster2021retrieval,santhanam2021rome} is still one of the most commonly used approaches. 
There are two main forms of human evaluation: (1) scoring, where human annotators rate the hallucination level in a range; 
and (2) comparing, where human annotators compare the output texts with baselines or ground-truth references \citep{sun2010mining}.

Multiple terminologies, such as
 \textit{faithfulness}~\citep{maynez2020faithfulness,cao2018faithful,chen2021improving,rashkin2021increasing,parikh2020totto,filippova2020controlled,rashkin2021increasing,tian2020sticking,su2021plan,xiao2021hallucination,zhou2021detecting}, 
 \textit{factual consistency}~\citep{cao2020factual,cao2021cliff,shen2021identifying, wu2021controllable, santhanam2021rome,chen2020logic2text}, \textit{fidelity}~\cite{chen2021confounded}, 
 \textit{factualness}\footnote{uses the source input as the “fact”. \label{fact}}~\cite{rebuffel2022controlling}, 
 \textit{factuality}$^4$~\citep{dong2020multi},
 or on the other hand, \textit{hallucination}~\citep{huang2020knowledge,shuster2021retrieval,santhanam2021rome,dziri2021neural,liu2021towards}, 
 \textit{fact contradicting}~\cite{nie2018operation}
  are used in the human evaluation of hallucination to rate whether the generated text is in accord with the source input.
 \citet{chen2021improving, nie2019simple} use finer-grained metrics for \textit{intrinsic hallucination} and \textit{extrinsic hallucination} separately. 
Moreover, there are some broad metrics, such as \textit{Correctness}~\citep{balakrishnan-etal-2019-constrained,bi2019incorporating,wang2021sketch,li2018ensure},
\textit{Accuracy}~\citep{yin2016neural,li2021addressing},
and \textit{Informativeness}~\citep{li-etal-2020-slot} 
considering both missing and additional contents (extrinsic hallucinations) compared to the input source.

\section{Hallucination Mitigation Methods}
\label{section:mitigation}

Common mitigation methods can be divided into two categories, in accordance with two main contributors of hallucinations: \textbf{Data-Related Methods}, and \textbf{Modeling and Inference Methods}.

\subsection{Data-Related Methods}
\subsubsection{Building a Faithful Dataset}
Considering that noisy data encourage hallucinations, constructing faithful datasets manually is an intuitive method, and there are various ways to build such datasets:
One way is employing annotators to write clean and faithful targets from scratch given the source~\cite{gardent2017creating, wen2015semantically}, which may lack diversity~\cite{parikh2020totto, gururangan2018annotation, poliak2018hypothesis}.
Another way is employing annotators to rewrite real sentences on the web~\cite{parikh2020totto}, 
or targets in the existing dataset~\cite{wang2019revisiting}. 
Basically, the revision strategy consists of three stages: 
(1) phrase trimming: removing phrases unsupported by the source in the exemplar sentence;
(2) decontextualization: resolving co-references and deleting phrases dependent on context;
(3) syntax modification: making the purified sentences flow smoothly.
Meanwhile, other works~\citep{honovich2021q,gabriel-etal-2021-go} leverage the model to generate data and instruct annotators to label whether these outputs contain hallucinations or not. 
While this approach is typically used to build diagnostic evaluation datasets, it has the potential to build faithful datasets.

\subsubsection{Cleaning Data Automatically}
In order to alleviate semantic noise issues, another approach is to find information that is irrelevant or contradictory to the input from the existing parallel corpus and then filter or correct the data. This approach is suitable for the case where there is a low or moderate level of noise in the original data~\citep{filippova2020controlled,nie2019simple}.

Some works~\cite{liu2021towards,shen2021identifying,Raunak2021} have dealt with the hallucination issue at the instance level by using a score for each source-reference pair and filtering out hallucinated ones.
 This corpus filtering method consists of several steps:
 (1) measuring the quality of the training samples in terms of hallucination utilizing the metrics described above; 
 (2) ranking these hallucination scores in descending order;
 (3) selecting and filtering out the untrustworthy samples at the bottom.
Instance-level scores can lead to a signal loss because divergences occur at the word level; i.e., parts of the target sentence are loyal to the source input, while others diverge~\citep{rebuffel2022controlling}.

Considering this issue, other works~\citep{nie2019simple,duvsek2019semantic} correct paired training samples, specifically the input data, according to the references. This method is mainly applied in the data-to-text task because structured data are easier to correcte than utterances.
This method consists of two steps: (1) utilizing a model to parse the meaning representation (MR), such as attribute-value pairs, from original human textual references; (2) using the MR extracted from the reference to correct the input MR through slot matching.
This method will enhance the semantic consistency between input and output without abandoning a part of the dataset.




\subsubsection{Information Augmentation}
\label{subsec:augment information}
It is intuitive that augmenting the inputs with external information will obtain a better representation of the source. Because the external knowledge, explicit alignment, extra training data, etc., can improve the correlation between the source and target and help the model learn better task-related features. Consequently, a better semantic understanding helps alleviate the divergence from the source issue.
Examples of the augmented information include entity information~\cite{liu2021towards}, 
extracted relation triples from source document~\cite{huang2020knowledge, cao2018faithful} obtained by Fact Description Extraction, \ziwei{pre-executed operation results~\citep{nie2018operation},}
 synthetic data generated through replacement or perturbation ~\cite{chen2021improving, lee2018hallucinations},
retrieved external knowledge~\cite{shuster2021retrieval, zhu2021enhancing, gunel2020mind,bi2019incorporating,fan2019using}, and retrieved similar training samples~\cite{bialecki2012apache}.

These methods enforce a stronger alignment between inputs and outputs. 
However, they will bring challenges due to the gap between the original source and augmented information, such as the semantic gap between an ambiguous utterance and a distinct MR of structured data, 
and the format discrepancy between the structured knowledge graph and natural language.


\subsection{Modeling and Inference Methods}

\subsubsection{Architecture}

\paragraph{Encoder} 
\vspace{-0.5em} 
The encoder learns to encode a variable-length sequence from input text into a fixed-length vector representation. 
As we mentioned above in Section \ref{subsec:augment information}, 
\ziwei{hallucination appears when the models lack semantic interpretation over the input.}
Some works have modified the encoder architecture in order to make it more compatible with input \ziwei{and learn a better representation}.
For example, \citet{huang2020knowledge} and \citet{cao2018faithful} propose a dual encoder, consisting of a sequential document encoder and a structured graph encoder to deal with the additional knowledge.

\vspace{-0.5em} 
\paragraph{Attention} 
The attention mechanism is an integral component in neural networks that selectively concentrates on some parts of sequences while ignoring others based on dependencies~\citep{vaswani2017attention}.
In order to encourage the generator to pay more attention to the source, \citet{aralikatte2021focus} introduce a short circuit from the input document to the vocabulary distribution via source-conditioned bias.
\citet{krishna2021hurdles} employ sparse attention to improve the model‘s long-range dependencies in the hope of modeling more retrieved documents so as to mitigate the hallucination in the answer.
\citet{wu2021controllable} adopt inductive attention, which removes potentially uninformative attention links by injecting pre-established structural information to avoid hallucinations.

\vspace{-0.5em} 
\paragraph{Decoder}
The decoder is responsible for generating the final output in natural language given input representations~\citep{vaswani2017attention}. 
Several work modified the decoder structures to mitigate hallucination, such as 
the multi-branch decoder~\cite{rebuffel2022controlling}, 
uncertainty-aware decoder~\cite{xiao2021hallucination},
dual decoder, consisting of a sequential decoder and a tree-based decoder~\cite{song2020joint},
and constrained decoder with lexical or structural limitations~\cite{balakrishnan-etal-2019-constrained}.
\yeon{Based on the observation that the ``randomness'' from sampling-based decoding, especially near the end of sentences, can lead to hallucination, \citep{lee2022factuality} propose to iteratively reduce the ``randomness'' through time. }
These decoders improve the possibility of faithful tokens while reducing the possibility of hallucinatory ones during inference by figuring out the implicit discrepancy and dependency between tokens or restricted by explicit constraints.
\ziwei{Since such decoders may have more difficulty generating fluent or diverse text, there is a balance to be struck between them.}

\subsubsection{Training}
\begin{figure*}[!t]
 \centering
 \includegraphics[width=0.9\linewidth]{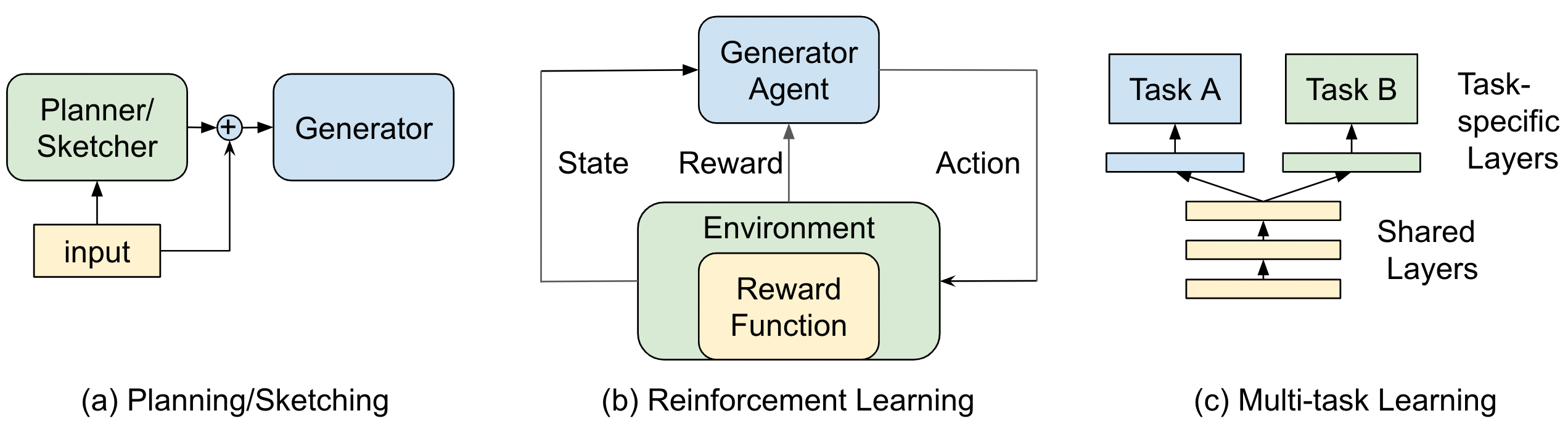}
  \caption{\ziwei{The frameworks of training methods.}}
  \label{fig:modeling}
    \vspace{-1em}
\end{figure*}

\paragraph{Planning/Sketching}
Planning is a common method to control and restrict what the model generates by informing the content and its order\ziwei{~\cite{puduppully2019data}}. Planning can be a separate step in a two-step generator\ziwei{~\cite{liu2021towards, su2021plan, chen2021improving,wang2021sketch,puduppully2021data}}\ziwei{, which is prone to progressive amplification of the hallucination problem.}
Or be injected into the end-to-end model during generation~\cite{xu2021agggen}.
 Sketching has a similar function to planning, and can also be adopted for handling hallucinations~\cite{wang2021sketch}. The difference is that the skeleton is treated as a part of the final generated text.
 \ziwei{While providing more controllability, such methods also need to strike a balance between faithfulness and diversity.}

\vspace{-0.5em} 
\paragraph{Reinforcement Learning (RL)}
As pointed out by \citet{ranzato2016sequence}, word-level maximum likelihood training leads to the problem of exposure bias.
Some works~\citep{huang2020knowledge, su2021plan, kong2019neural,mesgar-etal-2021-improving, li-etal-2020-slot} adopt RL to solve the hallucination problem, which utilizes different rewards to optimize the model.
The purpose of RL is for the agent to learn an optimal policy that maximizes the reward that accumulates from the environment~\citep{uc2021survey}.
The reward function is critical to RL and, if properly designed, it can provide training signals that help the model accomplish its goal of hallucination reduction.
For example, \citet{li-etal-2020-slot} propose a slot consistency reward which is the cardinality of the difference between generated template and the slot-value pairs extracted from input dialogue act. Improving the slot consistency can help reduce the hallucination phenomenon of missing or misplacing slot values in generated templates.
\citet{mesgar-etal-2021-improving} attain persona consistency sub-reward via an NLI model to reduce the hallucinations in personal facts.
\citet{huang2020knowledge} use a combination of ROUGE and the multiple-choice cloze score as the reward function to improve the faithfulness of summarization outputs. The cloze score is similar to the QA-based metric, measuring how well a QA model can address the questions by reading the generated summary (as context), where the questions are automatically constructed from the reference summary.
As the above examples show, some RL reward functions for mitigating hallucination are inspired by existing automatic evaluation metrics.
\ziwei{Although RL is challenging to learn and converge due to the extremely large search space, this method has the potential to obtain the best policy for the task without an oracle.}


\paragraph{Multi-task Learning}
\vspace{-0.5em} 
Multi-task learning is also utilized for handling hallucinations in different NLG tasks. 
In this training paradigm, a shared model is trained on multiple tasks simultaneously to learn the commonalities of the tasks. The hallucination problem may be derived from the reliance of the training process on a single dataset, leading to the fact that the model fails to learn the actual task features. By adding proper additional tasks along with the target task during training, the model can suffer less from the hallucination problem.
For example, \citet{weng2020towards} and \citet{garg2019jointly} incorporate a word alignment task into the translation model to improve the alignment accuracy between the input and output, and thus faithfulness.
\citet{li2018ensure} combine an entailment task with abstractive summarization to encourage models to generate summaries entailed by and faithful to the source.
\citet{li2021addressing} incorporate rationale extraction and the answer generation, which allows more confident and correct answers and reduces the hallucination problem.
The Multi-task approach has several advantages, such as data efficiency improvement, overfitting reduction, and fast learning. It is crucial to choose which tasks should be learned jointly, and learning multiple tasks simultaneously presents new challenges of design and optimization~\cite{crawshaw2020multi}.

\vspace{-0.5em} 
\paragraph{Controllable Generation}
Current works treat the hallucination level as a controllable attribute in order to remain the hallucination in outputs at a low level. 
Controllable generation techniques \ziwei{such as controlled re-sampling~\cite{rashkin2021increasing}, control codes that can be provided manually~\cite{rashkin2021increasing, filippova2020controlled, wu2021controllable}, or predicted automatically~\citep{wu2021controllable}} are leveraged to improve faithfulness.
\ziwei{This method may require some annotated datasets for training.}
Considering that hallucination is not necessarily harmful and may bring some benefits, controllable methods can be further adapted to change the degree of hallucination to meet the demands of different real-world applications. 

Other general training methods such as regularization~\cite{lee2018hallucinations, muller2020domain,kang2020improved} and loss reconstruction~\cite{wang2020towards,Li2021_TERM,wang2020exposure} have also been proposed to tackle the hallucination problem.


\subsubsection{Post-Processing} 
Post-processing methods can correct hallucinations in the output, and this standalone task requires less training data. Especially for noisy datasets where a large proportion of the ground truth references suffer from hallucinations, modeling correction is a competitive choice to handle the hallucination problem~\citep{chen2021improving}. 
\citet{chen2021improving,dong2020multi,cao2020factual}, and \citet{dziri2021neural} follow a generate-then-refine strategy. 
\ziwei{While the post-processing correction step tends to result in ungrammatical texts, this method} allows researchers to utilise SOTA models which perform best in respect of other attributes, such as fluency, and then correct the results specifically for faithfulness by using small amounts of automatically generated training data. 



\section{Future Directions}
\label{section:futurework}
Many studies have been conducted to tackle the hallucination problem in NLG and its downstream tasks. 
As mentioned above, we have discussed common metrics and mitigation methods to advance research in these fields.
From a broader perspective, we wish to point out open challenges and potential directions in regard to \textbf{metric} and \textbf{mitigation method}.

\subsection{Future Directions in Metrics Design}
\paragraph{Fine-grained Metrics}
Most of the existing hallucination metrics measure intrinsic and extrinsic hallucinations together as a unified metric. However, it is common for a single generation to have both types and a number of hallucinatory sub-strings. Fine-grained metrics that can distinguish between the two types of hallucinations will provide richer insight to researchers.

In order to implement a fine-graded metric, the first step would be to identify the exact location of the hallucinatory sub-strings correctly. However, some metrics such as those that are QA-based cannot identify the individual hallucinatory sub-strings. Improvements in this aspect would help improve the quality and explainability of the metrics. 
The next step would be to categorize the detected hallucinatory sub-strings. The hallucinatory sub-string will be intrinsic if it is wrong or nonsensical, and extrinsic if it is non-existing in the source context. Future work that explores an automatic method of categorization would be beneficial. 

\paragraph{Fact-Checking}
The factual verification of extrinsic hallucinations requires fact-checking against world knowledge, which can be time consuming and laborious.
Leveraging an automatic fact-checking system for extrinsic hallucination verification is, thus, other future work that requires attention.
Fact-checking consists of the knowledge evidence selection and claim verification sub-tasks, and the following are the remaining challenges associated with each sub-task. 

The main research problem associated with the evidence selection sub-task is how to retrieve evidence from the \textit{world} knowledge. Most of the literature leverages Wikipedia as the knowledge source~\cite{thorne-etal-2018-fever,yoneda2018ucl,lee2020language}, which is only a small part of world knowledge. Other literature attempts to use the whole web as the knowledge source
~\cite{Etzioni2008, magdy2010web}. However, this method leads to another research problem -- ``how to ensure the trustworthiness of the information we use from the web''~\cite{ginsca2015credibility}. Source-level methods that leverages the meta-information of the web source (e.g., web traffic, PageRank or URL structure) have been proposed to deal with this trustworthiness issue~\cite{popat2016credibility, baly2018predicting, popat2018declare}.
Addressing the aforementioned issues to allow evidence selection against world knowledge will be an important future research direction.

For the verification subtask, verification models perform relatively well if given correct evidence~\cite{leeimproving}. However, it has been shown that verification models are prone to adversarial attacks and are not robust to negation, numerical or comparative words~\cite{thorne-etal-2019-evaluating}.
Improving this weakness of verification models would also be crucial because the factuality of a sentence can easily be changed by small word changes (i.e., changes in negations, numbers, and entities).

\paragraph{Generalization}
Although we can see that the source and output text of different tasks are in various forms,
investigating their relationship and common ground and proposing general metrics to evaluate hallucinations are worth exploring. Task-agnostic metrics with cross-domain robustness could help the research community to build a unified benchmark.
It is also important and meaningful to build open-source platforms to collaborate and standardize the evaluation metrics for NLG tasks.

\paragraph{Incorporation of Human Cognitive Perspective}
A good automatic metric should correlate with human evaluation. 
Humans are sensitive to different types of information. For instance, proper nouns are usually more important than pronouns in the generated text. Mistakes concerning named entities are striking to human users, but automatic metrics treat them equally if not properly designed. In order to address this issue, new metrics should be designed from the human cognitive perspective. 
The human ability to recognize salient information and filter the rest is evident in scenarios where the most important facts need to be determined and assessed. For instance, when signing an agreement, a prospective employee naturally skims the document to look at the entries with numbers first. In this way, humans classify what they believe is crucial. 

Automatic check-worthy detection has the potential to be applied to improve the correlation with human judgement.
Implementing the automatic human-like judgment mentioned above can further mitigate hallucination and improve NLG systems.




\subsection{Future Directions in Mitigation Methods}
\paragraph{General and robust data pre-processing approaches} Since the data format varies between downstream tasks, there is still a gap for data processing methods between tasks, and currently, no universal method is effective for all NLG tasks~\cite{li2021data}.
Data pre-processing might result in grammatical errors or semantic transformation between the original and processed data, which can negatively affect the performance of generation.
Therefore, we believe that general and robust data pre-processing methods can help mitigate the hallucinations in NLG. 

\paragraph{Hallucinations in numerals} Most existing mitigation methods do not focus on the hallucination of numerals. However, the correctness of numerals in generated text, such as date, quantities and scalars are important for readers~\citep{zhao2020reducing,thawani2021representing, zhang2020language}. 
For example, given the source document ``\textit{The optimal oxygen saturation ($SpO_2$) in adults with COVID-19 who are receiving supplemental oxygen is unknown. However, a target $SpO_2$ of 92\% to 96\% seems logical, considering that indirect evidence from patients without COVID-19 suggests that an $SpO_2$ of <92\% or >96\% may be harmful.}~\footnote{\url{https://www.covid19treatmentguidelines.nih.gov/management/critical-care/oxygenation-and-ventilation/}}'', the summary ``\textit{The target oxygen saturation range for patients with COVID-19 is 82–86\%.}'' includes wrong numbers, which could be fatal.
Currently, some works~\cite{nie2019simple,thawani2021representing, zhang2020language} point out that using commonsense knowledge can help to gain better numeral representation. And \citet{zhao2020reducing} alleviate numeral hallucinations by re-ranking candidate-generated summaries based on the verification score of quantity entities.
Therefore, we believe that explicitly modeling numerals to mitigate hallucinations is a potential direction. 

\paragraph{Extrinsic Hallucination Mitigation} Though many works on mitigating hallucinations have been published, most do not distinguish between intrinsic and extrinsic hallucination. Moreover, the main research focus has been on dealing with intrinsic hallucination, while extrinsic hallucination has been somewhat overlooked as it is more challenging to reduce~\citep{huang2021factual}. Therefore, we believe it is worth exploring different mitigation methods for intrinsic and extrinsic hallucinations, and relevant methods in fact-checking can be potentially used for this purpose.

\paragraph{Hallucination in long text} 
Many tasks in NLG require the model to process long input texts, such as multi-document summarization and generative question answering. We think adopting existing approaches to a Longformer \citep{beltagy2020longformer}-based model could help encode long inputs. Meanwhile, part of dialogue systems need to generate long output text, in which the latter part may contradict history generation. Therefore, reducing self-contradiction is also an important future direction.

\paragraph{Reasoning} Misunderstanding facts in the source context will lead to intrinsic hallucination and errors. To help models understand the facts correctly requires reasoning over the input table or text. Moreover, if the generated text can be reasoned backwards to the source, we can assume it is faithful. There are some reasoning works in the area of dialogue~\cite{cui2020mutual, ghosal2021cider,wang2021fcm}, but few in reducing hallucinations. Moreover, tasks with quantities, such as logical table-to-text generation, require numerical reasoning. Therefore, adding reasoning ability to the hallucination mitigation methods is also an interesting future direction.

\paragraph{Controllability}
Controllability means the ability of models to control the level of hallucination and strike a balance between faithfulness and diversity~\citep{rohrbach2018object,dziri2021neural}. As mentioned in Section~\ref{section:contributors}, it is acceptable for chit-chat models to generate a certain level of hallucinatory content as long as it is factual. Meanwhile, for the abstractive summarization task, there is no agreement in the research community about whether factual hallucinations are desirable or not~\cite{maynez2020faithfulness}.
Therefore, we believe controllability merits attention when exploring hallucination mitigation methods.

\section{Hallucination in Abstractive Summarization}
\label{section:summarization}

Abstractive summarization aims to extract essential information from source documents and to generate short, concise, and readable summaries~\cite{yu2021adaptsum}. Neural networks have achieved remarkable results on abstractive summarization. However, \citet{maynez2020faithfulness} observe that neural abstractive summarization models are likely to generate hallucinatory content that is unfaithful to the source document. \citet{falke2019ranking} analyze three recent abstractive summarization systems and show that 25\% of the summaries generated from state-of-the-art models have hallucinated content. In addition, \citet{zhou2021detecting} mention that even if a summary contains a large amount of hallucinatory content, it can achieve a high ROUGE~\cite{lin2004rouge} score. This has encouraged researchers to actively devise ways to improve the evaluation of abstractive summarization, especially from the hallucination perspective.

In this section, we review the current progress in automatic evaluation and the mitigation of hallucination, and list the remaining challenges for future work. In addition, it is worth mentioning that researchers have used various terms to describe the hallucination phenomenon, such as faithfulness, factual errors, and factual consistency, and we will use the original terms from their papers in the remainder of this section.

\subsection{Hallucination Definition in Abstractive Summarization}

The definition of hallucination in abstractive summarization follows that in Section~\ref{section:definition}. Specifically, we adopt the definition from \citep{maynez2020faithfulness}: given a document and its abstractive summary, a summary is hallucinated if it has any spans not supported by the input document. 
Once again, intrinsic hallucination refers to output content that contradicts the source, while extrinsic hallucination refers to output content that the source cannot verify.
For instance, in Table~\ref{Table: example}, given the input article shown in the caption,
an example of intrinsic hallucination is ``\textit{The Ebola vaccine was rejected by the FDA in 2019,}'' because this statement contradicts the given content ``\textit{The first vaccine for Ebola was approved by the FDA in 2019 in the US}''.
And an example of extrinsic hallucination is ``\textit{China has already started clinical trials of the COVID-19 vaccine,}'' because this statement is not mentioned in the given content. We can neither find evidence of it from the input article nor assert that it is wrong.

\citet{pagnoni2021understanding} define fine-grained types of factual errors in summaries. 
As mentioned in \ref{subsec:terminology}, since the ``fact'' here refers to source knowledge, ``factual error'' can be treated as hallucination, and we can adopt this classification as a sub-type of hallucination.
They establish three categories as semantic frame error, discourse error, and content verifiability error. 


\subsection{Hallucination Metrics in Abstractive Summarization}
Existing metrics for hallucination in abstractive summarization are mainly model-based. Following~\citep{huang2021factual}, we divide the hallucination metrics into two categories: (1) unsupervised metrics and (2) semi-supervised metrics.
Note that existing hallucination metrics evaluate both intrinsic and extrinsic hallucinations together in one metric because it is difficult to automatically distinguish between them.

\subsubsection{Unsupervised Metrics} 
Given that hallucination is a newly emerging problem, there are only a few hallucination-related datasets. Therefore, researchers have proposed to adopt other datasets to build unsupervised hallucination metrics. There are three types of such unsupervised metrics: (1) information extraction (IE)-based metrics, (2) natural language inferencing (NLI)-based metrics, (3) question answering (QA)-based metrics.

\paragraph{IE-based Metrics} As mentioned in Section~\ref{section:metric}, IE-based metrics leverage IE models to extract knowledge as relation tuples (\textit{subject, relation, object}) from both the generation and knowledge source to analyze the factual accuracy of the generation~\cite{goodrich2019assessing}. However, IE models are not 100\% reliable yet (making errors in the identification of the relation tuples). Therefore, \citet{nan2021entity} propose an entity-based metric relying on the Named-Entity Recognition model, which is relatively more robust. Their metric builds on the assumption that there will be a different set of named entities in the gold and generated summary if there exists hallucination.


\paragraph{NLI-based Metrics} As mentioned in Section~\ref{section:metric}, the NLI-model (textual entailment model) can be utilized to measure hallucination based on the assumption that a faithful summary will be entailed by the gold source. 
However, \citet{falke2019ranking} discover that models trained on NLI datasets can not transfer well to abstractive summarization tasks, degrading the reliability of NLI-based hallucination metrics. To improve NLI models for hallucination evaluation, they release collected annotations as additional test data. Other efforts have also been made to further improve NLI models.  \citet{mishra2021looking} find that the low performance of NLI-based metrics is mainly caused by the length of the premises in NLI datasets being shorter than the source documents in abstractive summarization. Thus, the authors propose to convert multiple-choice reading comprehension datasets into long premise NLI datasets automatically. The results indicate that long-premise NLI datasets help the model achieve a higher performance than the original NLI datasets. In addition, \citet{laban2021summac} introduce a simple but efficient method called SUMMAC$_{Conv}$ by applying NLI models to sentence units that are segmented from documents. The performance of their model is better than applying NLI models to the whole document. 

\paragraph{QA-based Metrics}
QA-based metrics measure the knowledge overlap or consistency between summaries and the source documents based on the intuition that QA models will achieve similar answers if the summaries are factually consistent with the source documents. QA-based metrics such as FEQA~\citep{durmus2020feqa}, QAGS~\citep{wang2020asking}\ziwei{, and QuestEval~\citep{scialom2021questeval} follow} three steps to obtain a final score: (1) a QG model generates questions from the summaries, (2) a QA model obtains answers from the source documents, and (3) calculate the score by comparing the set of answers from source documents and the set of answers from summaries.
The results show that \ziwei{these reference-free metrics} have substantially higher correlations with human judgments of faithfulness than the baseline metrics. 
\citet{gabriel-etal-2021-go} further analyze the FEQA and find that the effectiveness of QA-based metrics depends on the question. They also provide a meta-evaluation framework that includes QA metrics.

\subsubsection{Semi-Supervised Metrics}
Semi-supervised metrics are trained on the synthetic data generated from summarization datasets. Trained on these task-specific corpora, models can judge whether the generated summaries are hallucinatory.
\citet{kryscinski2020evaluating} propose a weakly supervised model named FactCC for evaluating factual consistency. The model is trained jointly for three tasks: (1) checking whether the synthetic sentences remain factually consistent, (2) extracting supporting spans in the source documents, and (3) extracting inconsistent spans in the summaries, if any exist.
They transfer this model to check whether the summaries generated from summarization models are factually consistent. Results show that the performance of their FactCC model surpasses the classifiers trained on the MNLI or FEVER datasets. \citet{zhou2021detecting} introduce a method to fine-tune a pre-trained language model on synthetic data with automatically inserted hallucinations in order to detect the hallucinatory content in summaries. The model can classify whether spans in the machine-generated summaries are faithful to the article. This method shows higher correlations with human factual consistency evaluation than the baselines.

\subsection{Hallucination Mitigation in Abstractive Summarization}
Recently, many approaches have been proposed to reduce the hallucination phenomenon in abstractive summarization.

\subsubsection{Architecture Method.}
Seq-to-seq \citep{sutskever2014sequence} models are widely used and achieve state-of-the-art performance in abstractive summarization. Researchers have made modifications to the architecture design of the seq-to-seq models to reduce hallucinated content in the summaries. We describe various efforts made to improve the encoder, decoder, or both the encoder and decoder of the seq-to-seq models.

\paragraph{Encoder} \citet{zhu2021enhancing} propose to use an explicit graph neural network (GNN) to encode the fact tuples extracted from source documents. In addition to an explicit graph encoder, \citet{huang2020knowledge} further design a multiple-choice cloze test reward to encourage the model to better understand entity interactions. Moreover, \citet{gunel2020mind} use external knowledge from Wikipedia to make knowledge embeddings, which the results show improve factual consistency.

\paragraph{Decoder} \citet{song2020joint} present the incorporation of a sequential decoder with a tree-based decoder to generate a summary sentence and its syntactic parse. This joint generation is performed improve faithfulness. \citet{aralikatte2021focus} introduce the Focus Attention Mechanism, which encourages decoders to generate tokens similar or topical to the source documents. The results on the BBC extreme summarization task show that models augmented with the Focus Attention Mechanism generate more faithful summaries.

\paragraph{Encoder-decoder} \citet{cao2018faithful} extract fact descriptions from the source text and apply a dual-attention seq-to-seq framework to force the summaries to be conditioned on both source documents and the extracted fact descriptions. \citet{li2018ensure} propose an entailment-aware encoder and decoder with multi-task learning which incorporates the entailment knowledge into abstractive summarization models.
 
\subsubsection{Training Method} 
Aside from architecture modification, some works improved the training approach to reduce hallucination. \citet{cao2021cliff} introduce a contrastive learning method to train summarization models. The positive training data are reference summaries, while the negative training data are automatically generated hallucinatory summaries, and the contrastive learning system is trained to distinguish between them. In the dialogue summarization field, \citet{tang2021confit} propose another contrastive fine-tuning strategy, named CONFIT, that can improve the factual consistency and overall quality of summaries.

\subsubsection{Post-Processing Method} Some works carry out post-editing to reduce the hallucination of the model-generated summaries, which are viewed as draft summaries. \citet{dong2020multi} propose SpanFact, a pair of factual correction models that use knowledge learned from QA models to correct the spans in the generated summaries. Similar to SpanFact, \citet{cao2020factual} introduce a post-editing corrector module to identify and correct hallucinatory content in generated summaries. The corrector module is trained on synthetic data which are created by adding a series of heuristic transformations to reference summaries.
\citet{zhao2020reducing} present HERMAN, a system that learns to recognize quantities (dates, amounts of money, etc.) in the generated summary and verify their factual consistency with the source text. According to the quantity hallucination score, the system chooses the most faithful summary where the source text supports its quantity terms from the candidate-generated summaries. \citet{chen2021improving} introduce a contrast candidate generation and selection system to do post-processing. The contrast candidate generation model replaces the named entities in the generated summaries with ones present in the source documents, and the contrast candidate selection model will select the best candidate as the final output summary.

\subsection{Future Directions in Abstractive Summarization}
\paragraph{Factual Hallucination Evaluation} Factual hallucinations contain information not found in source content, though it is factually correct. In the summarization task, this kind of hallucination could lead to better summaries. However, there is little work focused on evaluating factual hallucination explicitly. Fact-checking approaches could be potentially used in this regard.

\paragraph{Extrinsic Hallucination Mitigation} There has been little research on extrinsic hallucinations as it is more challenging to detect and mitigate content based on world knowledge. We believe it is worth exploring extrinsic hallucination in terms of evaluation metrics and mitigation methods.

\paragraph{Hallucination in Dialogue Summarization} In conversational data, the discourse relations between utterances and co-references between speakers are more complicated than from, say, news articles. For example, \citet{zhong2021qmsum} show that 74\% of samples in the QMSum dataset consist of inconsistent facts. We believe exploring the hallucination issue in dialogue summarization is an important and special component of research into hallucination in abstractive summarization.

\section{Hallucination in Dialogue Generation}
\label{section:dialogue}

Dialogue generation is an NLG task that automatically generates responses according to user utterances. The generated responses are required to be fluent, coherent, and consistent with the dialogue history. The dialogue generation task can be divided into two sub-tasks: (1) task-oriented dialogue generation; (2) open-domain dialogue generation. A task-oriented dialogue system aims to complete a certain task according to a user query in a specific domain, such as restaurant booking, hotel recommendation, and calendar checking. Meanwhile, an open-domain dialogue system aims to establish a multi-turn, long-term conversation with users while providing the users with an engaging experience.

\subsection{Hallucination Definition in Dialogue Generation}
The hallucination problem also exists in the dialogue generation task. It is important to note that a dialogue system is expected either to provide the user with the required information or to provide an engaging response without repeating utterances from the dialogue history. Thus, the tolerance for producing proper ``hallucination'' from the dialogue history is relatively higher.

The definition of hallucination in this task can be adopted from the general definition as follows: (1) \textbf{Intrinsic hallucination}: the generated response is contradictory to the dialogue history or the external knowledge sentences. In the examples of intrinsic hallucination shown in Table~\ref{Table: example}, we can verify that the output contradicts the inputs: In one example, the input is a ``\textit{moderate}'' price range, but the model mistakenly generates a sentence with a ``\textit{high}'' price range. In another case, the confusion of the names ``\textit{Roger Federer}'' and ``\textit{Rafael Nadal}'' causes the output generation of ``\textit{Roger Nadal}''. (2) \textbf{Extrinsic hallucination}: the generated response is hard to verify with the dialogue history or the external knowledge sentences. Responses with extrinsic hallucination are impossible to verify with the given inputs. ``\textit{Pickwick hotel}'' might be ``\textit{in san diego}'', and Djokovic may have been ``\textit{in the top ten singles players of the world}''. However, we do not have enough information to check the truth of these statements.

In the following sections, the hallucination problem in open-domain and task-oriented dialogue generation tasks will be separately discussed according to the their natures.

\subsection{Open-domain Dialogue Generation}
While the term ``hallucination'' seems to have newly emerged in the NLP field, a related behavior, ``inconsistency'', of neural models has been widely discussed. This behavior has been pointed out as a shortcoming of generation-based approaches for open-domain chatbots~\citep{huang2020challenges, ma2020survey, roller2020open}. 
Two possible types of inconsistency occur in open-domain dialogue generation: (1) inconsistency among the system utterances, such as when the system contradicts its previous utterance; (2) inconsistency with an external source, such as factually incorrect utterances. Whereas the first type is described using the term "inconsistency"~\citep{welleck-etal-2019-dialogue, li-etal-2020-dont, zhang2021dscore} or "incoherence"~\citep{dziri-etal-2019-evaluating, beyer-etal-2021-incoherence}, some have recently started to call the second type "hallucination"~\citep{roller2021recipes, mielke2020linguistic}.  Self-inconsistency can be considered as an intrinsic hallucination problem, while the external inconsistency involves both intrinsic and extrinsic hallucinations, depending on the reference source. 

As mentioned earlier, a certain level of hallucination may be acceptable in open-domain chit-chat as long as it does not involve severe factual issues. Moreoever, it is almost impossible to verify factual correctness since the system usually lacks a connection to external resources. With the introduction of knowledge-grounded dialogue tasks~\citep{zhou2018dataset, dinan2018wizard}, which provide an external reference, however, there has been more active discussion of hallucination in open-domain dialogue generation.


\subsubsection{Self-Consistency}
In end-to-end generative open-domain dialogue systems, the inconsistency among system utterances has been pointed out as the bottleneck to human-level performance ~\citep{vinyals2015neural}. We often observe an inconsistency in the answers to semantically similar yet not identical questions. For example, a system may answer the questions of ``What is your name?'' and ``May I ask your name?" with different responses. Persona consistency has been the center of attention~\citep{li2016persona, zhang2018personalizing} and it is one of the most obvious cases of self-contradiction regarding the character of the dialogue system. "Persona" is defined as the character that a dialogue system plays during a conversation, and can be composed of identity, language behavior, and an interaction style~\citep{li2016persona}. While some works has set their objective as teaching models to utilize speaker-level embeddings~\citep{li2016persona, madotto2019personalizing}, others condition generation with a set of descriptions about a persona, which we will discuss in detail in the next section.

\subsubsection{External Consistency}
Besides self-consistency, an open-domain dialogue system should also generate persona-consistent and informative responses corresponding so as to user utterances to further engage with the user during conversation. In this process, an external resource containing explicit persona information or world knowledge is introduced into the system to assist the model generation process.

The PersonaChat datasets~\cite{zhang2018personalizing, dinan2020second} have accelerated research into persona consistency~\citep{hancock2019learning, kulikov2019importance, mazare2018training, yavuz2019deepcopy, zemlyanskiy2018aiming, wolf2019transfertransfo,
zhang-etal-2020-dialogpt}. In PersonaChat datasets, each conversation has persona descriptions such as ``I like to ski'' or ``I am a high school teacher'' attached. By conditioning the response generation on the persona description, a chit-chat model is expected to acquire an ability to generate a more persona-consistent response. Lately, the application of NLI methods~\citep{li-etal-2020-dont, song2020generating} or reinforcement learning frameworks~\citep{mesgar-etal-2021-improving} have been investigated. Although these methods conditioned on the PersonaChat datasets have been successful, further investigation of approaches that do not rely on a given set of persona descriptions is necessary because such descriptions are not always available, and covering every aspect of a persona with them is impossible. 

In addition to PersonaChat-related research, the knowledge-grounded dialogue (KGD) task in the open-domain requires the model to generate informative responses with the help of an external knowledge graph (KG) or knowledge corpus~\cite{zhou2018dataset,dinan2018wizard}. Hallucination in conversations, which is also considered as a factual consistency problem, has raised much research interest recently~\cite{dziri2021neural,shuster2021retrieval,rashkin2021increasing,santhanam2021rome}. Here, we continue to split the hallucination problem in the KGD task into intrinsic hallucination and extrinsic hallucination. Most of the KGD works tackle the hallucination problem when responses contain information that contradicts (intrinsic) or cannot be found in the provided knowledge input (extrinsic). 
Since world knowledge is enormous and ever-changing, the extrinsic hallucination may be factual but hard to verify. \citet{dziri2021neural} further adopt the same definition of hallucination as mentioned above to the knowledge graph-grounded dialogue task, where intrinsic hallucination indicates the case of misusing either the subject or object of the knowledge triple; 
and extrinsic hallucination indicates that there is no corresponding valid knowledge triple in the gold reference knowledge.
Recently, there have been some attempts to generate informative responses without explicit knowledge inputs, but with the help of the implicit knowledge inside large pre-trained language models instead~\cite{xu2021retrieval,zhou2021think} during the inference time. Under this setting, the study of extrinsic hallucination is of great value but still poorly investigated.

\subsubsection{Hallucination Metrics}
For generation-based dialogue systems, especially open-domain chatbots, the hallucination evaluation method remains an open problem~\citep{roller2020open}. As of now, there is no standard metric. Therefore, chatbots are usually evaluated by humans on factual consistency or factual correctness~\citep{wu2021controllable,santhanam2021rome}. We also introduce some automatic statistical and model-based metrics as a reference, which will be described in more detail below.

\paragraph{Variants of F1 Metrics} 
\textbf{Knowledge F1 (KF1)} measures the overlap between the generated responses and the gold knowledge sentences to which the human referred for conversation during dataset collection~\cite{shuster2021retrieval}. %
KF1 attempts to capture whether a model can generate knowledgable responses by correctly utilizing the relevant knowledge. %
KF1 is only available for datasets with labeled ground-truth knowledge. \citet{shuster2021retrieval} further propose \textbf{Rare F1 (RF1)}, which only considers the infrequent words in the dataset when calculating F1 to avoid influence from the common uni-grams. The authors define an infrequent word as being in the lower half of the cumulative frequency distribution of the reference corpus.

\paragraph{Model-based Metric} 
\yan{Natural language has its natural on the flexibility of the surface forms with the same semantics, so overlap-based metrics cannot provide the comprehensive evaluation.} 
Recently, several works have proposed evaluation metrics for measuring consistency, such as using natural language inference (NLI)~\citep{welleck-etal-2019-dialogue, dziri-etal-2019-evaluating}, training learnable evaluation metrics~\citep{zhang2021dscore}, or releasing an additional test set for coherence~\citep{beyer-etal-2021-incoherence}.
\yan{These methods are more flexible and supports the generated responses with different surface forms.} 
For the KGD task, \citet{dziri2021evaluating} propose the BEGIN benchmark, which consists of samples taken from \citet{dinan2018wizard} with additional human annotation and a new classification task extending the NLI paradigm. \citet{honovich2021q} present a trainable metric for the KGD task, which also applies NLI. It is also noteworthy that \citet{gupta2021dialfact} propose datasets that can benefit fact-checking systems specialized for dialogue systems.
The Conv-FEVER corpus~\citep{santhanam2021rome} is a factual consistency detection dataset, which was created by adapting the Wizard-of-Wikipedia dataset~\citep{dinan2018wizard}. It consists of both factually consistent and inconsistent responses and can be used to train a classifier to detect factually inconsistent responses with respect to the knowledge provided.

\subsubsection{Mitigation Methods}
The hallucination issue can be mitigated by data pre-processing, which includes introducing extra information into the data. \citet{shen2021identifying} propose a measurement based on seven attributes of the dialogue quality, including self-consistency. Based on this measurement, the untrustworthy samples which get lower scores are filtered out from the training set to improve the model performance in terms of self-consistency (i.e., intrinsic hallucination). 
\citet{shuster2021retrieval} conduct a comprehensive investigation on a retrieval-augmented KGD task where a retriever is introduced to the system for knowledge selection. The authors study several key problems, such as whether retrieval helps reduce hallucinations and how the generation should be augmented with the retrieved knowledge. The experimental results show that retrieval helps substantially in improving performance on KGD tasks and in reducing the hallucination in conversations without sacrificing conversational ability.

\citet{rashkin2021increasing} introduce a set of control codes and concatenate them with dialogue inputs to reduce the hallucination by forcing the model to be more aware of how the response relies on the knowledge evidence in the response generation.
Some researchers have also tried to reduce hallucinated responses during generation by improving dialogue modeling. \citet{wu2021controllable} apply inductive attention into transformer-based dialogue models, and potentially uninformative attention links are removed with respect to a piece of pre-established structural information between the dialogue context and the provided knowledge.
Instead of improving the dialogue response generation model itself, \citet{dziri2021neural} present a response refinement strategy with a token-level hallucination critic and entity-mention retriever, so that the original dialogue model is left without retraining. The former module is designed to label the hallucinated entity mentioned in the generated responses, while the retriever is trained to retrieve more faithful entities from the provided knowledge graph.
RHO \cite{ji-etal-2023-rho} is a framework that uses three mechanisms to tackle hallucinations, namely, local knowledge grounding, global knowledge grounding, and response re-ranking to tackle hallucinations in open-domain dialogues, and has been empirically shown to perform this.

\subsection{Task-oriented Dialogue Generation}
A task-oriented dialogue system is often composed of several modules: a natural language understanding (NLU) module, a dialogue manager (DM), and a natural language generation (NLG) module~\cite{gao2018neural, jurafsky2019speech}. Intrinsic hallucination can occur between the DM and NLG, where a dialogue act such as \texttt{recommend(NAME=}\textit{peninsula hotel}\texttt{, AREA=}\textit{tsim sha tsui}\texttt{)} is transformed into a natural language representation ``the hotel named \textit{peninsula hotel} is located in \textit{tsim sha tsui} area.''~\citep{balakrishnan-etal-2019-constrained, li-etal-2020-slot}. 

\subsubsection{Hallucination Metrics}
To evaluate hallucination, \citet{li-etal-2020-slot} and \citet{balakrishnan-etal-2019-constrained} combine traditional metrics such as the BLEU score and human evaluation as well as hallucination-specific automatic metrics. Following previous works such as~\citep{wen-etal-2015-stochastic, dusek-jurcicek-2016-sequence}, and \citep{tran-nguyen-2017-natural}, \citet{li-etal-2020-slot} use the slot error rate, which is computed by $(p+q)/N$, where $N$ represents the total number of slots extracted by another model in the dialogue act. Here, $p$ stands for the missing slots in the generated template, and $q$ is the number of redundant slots. 
On the other hand, \citet{balakrishnan-etal-2019-constrained} introduce a novel metric called the tree accuracy, which determines whether the prediction's tree structure is identical to that of the input meaning representations.

\subsubsection{Mitigation Methods}
While \citet{balakrishnan-etal-2019-constrained} propose to adopt tree-structured semantic representations and add constraints on decoding, \citet{li-etal-2020-slot} frame a reinforcement learning problem to which they apply a bootstrapping algorithm to sample training instances and then leverage a reward related to slot consistency. 
Recently, there has emerged another line of research in task-oriented dialogue, which is to build a single end-to-end system rather than connecting several modules (e.g., \citet{eric-manning-2017-copy, wu2018globaltolocal, madotto-etal-2018-mem2seq, madotto-etal-2020-learning}). As discussed in previous sections of this paper, there is potential for such end-to-end systems to produce extrinsic hallucinations, yet this remains less explored. For example, a model might generate a response with an entity that appears out of nowhere. In the example of hotel recommendation in Hong Kong given above, a model could generate a response such as ``the hotel named \textit{raffles hotel} is located in \textit{central} area,\footnote{Raffles Hotel is a hotel located in Downtown Core, Singapore.}'' which cannot be verified from the knowledge base of the system.

\subsection{Future Directions in Dialogue Generation}
\paragraph{Self-Contradiction in Dialogue Systems} 
One of the possible reasons for self-contradiction is that current dialogue systems tend to have a short memory of dialogue history~\citep{roller2020open}. 
Firstly, common dialogue datasets provide several turns of conversation, yet these are not long enough to assess a model's ability to deal with a long context. To overcome this, \citet{xu2021beyond} introduce a new dataset that consists of, on average, over 40 utterances per episode. Secondly, we often truncate dialogue history into fewer turns to fit into models such as Transformer-based architectures, which makes it difficult for a model to memorize the past. In addition to the works on dialogue summarization, e.g.,~\citet{gliwa-etal-2019-samsum}, it would be beneficial to apply other works which are aiming to grasp the longer context but do not focus on dialogue generation~\citep{beltagy2020longformer, zaheer2020big, zhao-etal-2021-ror-read}.

\paragraph{Fact-checking in dialogue systems} 
In addition to the factual consistency in responses from knowledge grounded dialogue systems, fact-checking is a future direction in dealing with the hallucination problem in dialogue systems~\citep{gupta2021dialfact}. Dialogue fact-checking involves verifiable claim detection, which is an important line in distinguishing hallucination-prone dialogue, and evidence retrieval from an external source. This fact-checking in the dialogue system could be utilized not only as an evaluation metric for facilitating factual consistency but also to model such a system.

\section{Hallucination in Generative Question Answering}
\label{section:QA}
Generative question answering (GQA) aims to generate an abstractive answer rather than extract an answer to a given question from provided passages~\cite{fan2019eli5,li2021addressing}. It is an important task since many of the everyday questions that humans deal with and pose to search engines require in-depth explanations~\cite{khashabi2021gooaq} (e.g., \textit{why/how..?}), and the answers are normally long and cannot be directly extracted from existing phrase spans. A GQA system can be integrated with a search engine~\cite{metzler2021rethinking} to empower more intelligent search or combined with a virtual conversation agent to enhance user experience. 

Normally, a GQA system involves searching an external knowledge source for information relevant to the question. Then it generates the answer based on the retrieved information~\cite{krishna2021hurdles}. In most cases, no single source (document) contains the answer, and multiple retrieved documents will be considered for answer generation. Those documents may contain redundant, complementary, or contradictory information. Thus, hallucination is common in the generated answers. 

The hallucination problem is one of the most important challenges in GQA. Since an essential goal of a GQA system is to provide factualy-correct answers given the question, hallucination in the answer will mislead the user and damage the system performance dramatically. 

\subsection{Hallucination Definition in GQA}
As a challenging yet under-explored task, there is no standard definition of hallucination in GQA. However, almost all the works on GQA~\cite{fan2019eli5,krishna2021hurdles, nakano2021webgpt,su2022read} involve a human evaluation process, in which the \textit{factual correctness} measuring the faithfulness of the generated answer can be seen as a measurement of the hallucination; i.e., the more faithful the answer is, the less hallucinated content it contains. The most recent such work ~\cite{li2021addressing} uses the term \textit{semantic drift}, which indicates how the answer drifts away from a correct one during generation, and this can also be seen as a specific definition of hallucination in GQA. 

In line with the general categorization of hallucination in Section ~\ref{subsec:category}, we give two concrete hallucination examples in GQA in Table ~\ref{Table: example}. The sources of both questions are Wikipedia web pages. For the first question, ``\textit{dow jones industrial average please?}'', the generated answer ``\textit{index of 30 major U.S. stock indexes}'' conflicts with the statement ``\textit{of 30 prominent companies listed on stock exchanges in the United States}'' from Wikipedia. So we categorize it as an {intrinsic hallucination}. For the second example, the sentences ``\textit{The definition of a Sadducee is a person who acts in a deceitful or duplicitous manner. An example of a Sadduceee is a politician who acts deceitfully in order to gain political power}'' in the generated answer can not be verified from the source documents; thus, we categorize it as an {extrinsic hallucination}.

\subsection{Hallucination-related Metrics in GQA} 
Currently, there is no automatic metric to evaluate hallucination in QGA specifically. While most works on GQA use automatic evaluation metrics such as ROUGE score and F1 to measure the quality of the answer, these N-gram overlap-based metrics are not a meaningful way to evaluate hallucination due to their poor correlation with human judgments, as indicated by \citet{krishna2021hurdles}. On the other hand, almost all the GQA-related work involves a human evaluation process as a complement to the automatic evaluation. Normally, human annotators will be asked to assign a score indicating the faithfulness of the answer, which can also be viewed as a measurement of the answer hallucination. However, the metrics obtained via human evaluation come normally from a small sample of the data.

Metrics such as \textit{semantic overlap} ~\cite{sellam2020bleurt}, a learned evaluation metric based on BERT that models human judgments, could be considered a better measurement of hallucination for GQA. Other metrics such as the \textit{factual correctness} can also be considered as a way to measure hallucination in GQA. \citet{zhang2020optimizing} propose to explicitly measure the factual correctness of a generated text against the reference by first extracting facts via an information extraction (IE) module. Then they define and measure the factual accuracy score to be the ratio of facts in the generation text equal to the corresponding facts in the reference.

\textit{Factual consistency}, which measures the faithfulness of the generated answer given its source documents, can be employed as another way to measure hallucination in GQA. \citet{durmus2020feqa, wang2020asking} propose an automatic QA-based metric to measure faithfulness in summary, leveraging the recent advances in machine reading comprehension. They first use a question generation model to construct question-answer pairs from the summary, and then a QA model is applied to extract short answer spans from the given source document for the question. The extracted answers that do not match the provided answers indicate unfaithful information in the summary. While these metrics were first proposed in summarization works, they can be easily adopted in generative QA to measure hallucinations in the generated long-form answer.

The most recent work on GQA by ~\citet{su2022read} proposed to estimate the faithfulness of the generated long-form answer via \textit{zero-shot short answer recall} on extractive QA datasets. They first generate long-form answers for questions from two extractive QA datasets Natural Questions(NQ)~\cite{kwiatkowski2019natural} and HotpotQA~\cite{yang2018hotpotqa}, both of which contains large-scale question-answer pairs, then they measure the ratio of golden short answer span contained in the generated long answer as an estimation of faithfulness of the generated long-answer. While the idea is similar to the factual consistency metric in summarization work~\cite{durmus2020feqa}, and also matches with our intuition to some extent, its correlation with human evaluation on faithfulness has not been verified.


\subsection{Hallucination Mitigation in GQA}
Unlike conditional text generation tasks such as summarization, or data-to-text generation, in which the source documents are provided and normally related to the target generation, the hallucination problem in GQA is more complicated. Generally speaking, it might come from two sources: 1) the incompetency of the retriever, which retrieves documents irrelevant to the answer, and 2) the \textit{intrinsic} and \textit{extrinsic} hallucination in the conditional generation model itself. Normally these two parts are interconnected and cause hallucinations in the answer.

Early works on GQA mostly tried to improve the faithfulness of the answer by investigating reliable external knowledge sources or incorporating multiple information sources.
\citet{yin2016neural} propose Neural Generative Question Answering (GENQA), an end-to-end model that generates answers to simple factoid questions based on the knowledge base, while \citet{bi2019incorporating} propose the Knowledge-Enriched Answer Generator (KEAG) to generate a natural answer by integrating facts from four different information sources, namely, questions, passages, vocabulary, and knowledge. 
\ziwei{Nevertheless, these methods rely on the existence of high-quality, relevant resources which are not easily available.}

Recent works focus more on the conditional generation model. \citet{fan2019using} construct a local knowledge graph for each question to compress the information and reduce redundancy from the retrieved documents, which can be viewed as an early trial to mitigate hallucination. \citet{li2021addressing} propose Rationale-Enriched Answer Generator (REAG), in which they add an extraction task to obtain the rationale for an answer at the encoding stage, and the decoder is expected to generate the answer based on both the extracted rationale and original input. The recent work~\citep{krishna2021hurdles} employs a Routing Transformer (RT), a sparse attention-based Transformer-based model that employs local attention and mini-batch k-means clustering for long-range dependence, as the answer generator in the hope of modeling more retrieved documents to mitigate the hallucination in the answer. \citet{su2022read} propose a framework named RBG (\textbf{r}ead \textbf{b}efore \textbf{g}enerate), to jointly models answer generation with machine reading. They augment the generation model with fine-grained, answer-related salient information predicted by the MRC module, to enhance answer faithfulness.
\ziwei{Such methods can exploit and utilize the information in the original input better, while they require the extra effort of building models to extract that information.}

Most recently,~\citet{lin2021truthfulqa} propose a benchmark, which comprises 817 questions that span 38 categories, to measure the truthfulness of a language model in the QA task. This work investigates the performances of GPT-3~\cite{brown2020language}, GPT-Neo/J~\cite{gpt-j}, GPT-2~\cite{radford2019language} and a T5-based model~\cite{2020t5}. The results suggest that simply scaling up the model is less promising than fine-tuning it in terms of improving truthfulness since larger models are better at learning the training distribution from web data and thus tend to produce more imitative falsehoods. In another recent work, \citet{nakano2021webgpt} fine-tune GPT-3 to answer long-form questions with a web-browsing environment, which allows the model to navigate the web as well as use human feedback to optimize answer quality directly using imitation learning~\cite{10.1145/3054912}. 
\ziwei{While this method seems promising, it also hinges on how that feedback is processed.}

\subsection{Future Directions in GQA} 
While GQA is challenging yet under-explored, many possible directions could be explored to improve the answer quality and mitigate hallucination. 
First, better automatic evaluation metrics are needed to measure hallucination. The previously mentioned metrics, such as the semantic overlap between the generated answer and the ground-truth answer, the faithfulness of the generated answer, and factual consistency between the answer and the source documents, only consider one aspect of hallucination. Metrics that can consider all the factors related to hallucination (such as semantic overlap, faithfulness, or factual consistency) could be designed. 
Second, datasets with hallucination annotations should be proposed since none of the current GQA datasets have that information. Another possible direction to mitigate hallucination in the answer is improving the performance of the models. We need better retrieval models that retrieve relevant information according to queries and generation models that can synthesize more accurate answers from multi-source documents.

\section{Hallucination in Data-to-Text Generation}
\label{section:data2text}
Data-to-Text Generation is the task of generating natural language descriptions conditioned on structured data~\cite{kukich1983design, mckeown1992text}, such as tables~\cite{parikh2020totto, wiseman2017challenges}, database records~\cite{chisholm2017learning}, and knowledge graphs~\cite{gardent2017creating}.
Although this field has been recently boosted by neural text generation models, it is well known that these models are prone to hallucinations~\cite{wiseman2017challenges} because of the gap between structured data and text, which may cause semantic misunderstanding and erroneous correlation.
Moreover, the tolerance of hallucination is very low when this task is applied to the real world, such as in the case of patient information table description~\citep{thomson2020gold}, and analysis of experimental results tables in a scientific report.
Recent years have seen a growth of interest in hallucinations in Data-to-Text Generation, and researchers have proposed works from the aspect of evaluation and mitigation.



\subsection{Hallucination Definition in Data-to-Text Generation} 
The definition and categories of hallucination in Data-to-Text Generation follow the descriptions in Section \ref{section:definition}. We follow the general hallucination definition in this task:
(1) Intrinsic Hallucinations: the generated text contains information that is contradicted by the input data~\cite{nie2019simple}. For example, in Table \ref{Table: example}, ``\textit{The Houston Rockets (18-4)}'' uses the information ``\textit{[TEAM: Rockets, CITY:Houston, WIN:18, LOSS: 5]}'' in the source table. However, ``\textit{(18-4)}'' is contradicted by ``\textit{[LOSS: 5]}'' and it should be ``\textit{(18-5)}''.
(2) Extrinsic Hallucinations: the generated text contains extra information irrelevant to the input ~\cite{dhingra2019handling, nie2019simple}. For example, in Table \ref{Table: example}, ``\textit{Houston has won two straight games and six of their last seven.}'' is not mentioned in the source table~\cite{wang2019revisiting}.

\subsection{Hallucination Metrics in Data-to-Text Generation} 
\paragraph{Statistical}
PARENT~\cite{dhingra2019handling} measures the accuracy of table-to-text generation by aligning n-grams from the reference description \ziwei{$R$ and generated texts $G$ to the table $T$}. And it is the average F-score by combining the entailment precision and recall. 
\citet{wang2020towards} modify PARENT and denote this table-focused version as PARENT-T.  
Different from PARENT, which evaluates \ziwei{i-th} instance $(T_i, R_i, G_i)$, PARENT-T ignores the reference description R and evaluates each instance $(T_i, G_i)$.

\paragraph{IE-based}
\vspace{-0.5em} 
\citet{liu2021towards} estimate hallucination with two entity-centric metrics: table record coverage (the ratio of covered records in a table) and hallucinated ratio (the ratio of hallucinated entities in text).
This metric firstly uses entity recognition to extract the entities of input and generated output,
then aligns these entities by heuristic matching strategies,
and finally calculates the ratios of faithful and hallucinated entities separately.
Moreover, there are some general post-hoc IE-based metrics that could be applied to hallucination evaluation, such as Slot Error Rate (SER)~\citep{xu2021agggen}, 
Content Selection (CS), Relation Generation (RG), and Content Ordering (CO)~\cite{wiseman2017challenges, wang2019revisiting}.

\paragraph{QA-based}
\vspace{-0.5em} 
\ziwei{Data-QuestEval~\citep{rebuffel2021data} adapt QuestEval~\citep{scialom2021questeval} from summarization into data-to-text generation. 
First, a \textit{textual QG model} is trained on a textual QA dataset. 
For each sample (structured data, textual descriptions), the \textit{textual QG model} generates synthetic problems based on the descriptions.
The structured data, textual descriptions (answers), and synthetic questions make up a synthetic QG/QA dataset to train \textit{synthetic QA/QG models}.
Then, the \textit{synthetic QG} model generates questions based on the textual description to be evaluated. The \textit{synthetic QA} model then generates answers based on a synthetic question and the structured input data.
Finally, BERTScore~\citep{zhang2019bertscore} measures the similarity between the generated answer and description, indicating faithfulness.
}

\paragraph{NLI-based}
\vspace{-0.5em} 
\citet{dusek-kasner-2020-evaluating} recognize the textual entailment between the input data and the output text for both omissions and hallucinations with an NLI model. This work measures the semantic accuracy in two directions: check omissions by inferring whether the input fact is entailed by the generated text and check hallucinations by inferring the generated text from the input.

\paragraph{LM-based}
\vspace{-0.5em} 
\citet{filippova2020controlled, tian2020sticking} are based on the intuition that when an unconditional LM, only trained on the targets, gets a smaller loss than a conditional $LM_x$, trained on both sources and targets, the token is predicted unfaithfully. Thus, they calculate the ratio of hallucinated tokens to the total target length to measure the hallucination level.

\subsection{Hallucination Mitigation in Data-to-Text Generation}
\label{subsec:mitigation in data2text}
\paragraph{Data-Related Methods}
Several clean and faithful corpora are collected to tackle the challenges from data infidelity. 
TOTTO~\cite{parikh2020totto} is an open-domain faithful table-to-text dataset, where each sample includes a Wikipedia table with several highlighted cells and a description. 
To ensure that targets exclude hallucinations, the annotators revise existing Wikipedia candidate sentences and clear the parts unsupported by the table.  
Moreover, RotoWire-FG (Fact-Grounding)~\cite{wang2019revisiting} is a purified and enlarged and enriched version of RotoWire~\cite{wiseman2017challenges} generating NBA game summaries from score tables.
Annotators trim the hallucination part in target texts and extract the mapped table records as content plans to better align input tables and output summaries.

For data processing, \ziwei{OpAtt~\citep{nie2018operation} designs a gating mechanism and a quantization module for the symbolic operation to augment the record table with pre-calculated results.}
\citet{nie2019simple} utilize a language understanding module to improve the equivalence between the input MR and the reference utterance in the dataset.
They train an NLU model with an iterative relabeling procedure: 
First, they train the model on original data; parse the MR by model inference; train the model on new paired data with high confidence; and then repeat the above processes.
\citet{liu2021towards} select training instances based on faithfulness ranking. Finer-grained than the above instance-level method, \citet{rebuffel2022controlling} label tokens according to co-occurrence analysis and sentence structure through dependency parsing in the pre-processing step to explicate the correspondence between the input table and the text. 
\ziwei{Generally, the data-related methods are appropriate when the training dataset is noisy.}

\paragraph{Modeling and Inference Methods}
\vspace{-0.5em} 
Planning and skeleton generation are common methods to improve the faithfulness to the input in data-to-text tasks. 
\citet{liu2021towards} propose a two-step generator with a separate text planner augmented by auxiliary entity information.
The planner predicts the plausible content plan based on the input data. Then, given the above input data and the content plan, the sequence generator generates the text. 
Similarly, Plan-then-Generate~\cite{su2021plan} also consists of a content planner and a sequence generator.
In addition, this work adopts a structure-aware RL training to generate output text following the generated content plan faithfully. 
\ziwei{\citet{puduppully2021data} first induce a macro plan consisting of multiple sequences of entities and events from the input table and its corresponding multi-paragraph long document. The predicted macro plan then serves as the input to an encoder-decoder model for surface realization.}
SANA~\cite{wang2021sketch} is a skeleton-based two-stage model that includes skeleton generation to select key tokens from the source table and edit-based generation to produce texts via iterative insertion and deletion operations.
In contrast to the above two-step model using planning or skeleton, AGGGEN~\cite{xu2021agggen} is an end-to-end model that jointly learns to plan and generate at the same time.
This architecture with a Hidden Markov Model and Transformer encoder-decoder reintroduces explicit sentence planning stages into neural systems by aligning facts in the target text to input representations.

Other modeling methods have also been proposed to mitigate the hallucination problem.
Conjecturing that hallucinations can be caused by inattention to the source, \citet{tian2020sticking} propose a confidence score and a variational Bayes training framework to learn the score from data.  
\citet{wang2020towards} introduce a table-text optimal-transport matching loss and an embedding similarity loss to encourage faithfulness.
The hallucination degree can also be treated as a controllable factor in generating texts.
In \citet{filippova2020controlled}, the hallucination degree of each training sample is estimated and converted into a categorical value which is a part of the inputs as a controlled setting. 
This approach does not require the dismissal of any input or modification of the model structure.

To mitigate hallucinations at the inference step,~\citet{rebuffel2022controlling} propose a Multi-Branch Decoder that leverages word-level alignment labels between the input table and paired text to learn the relevant parts of the training instance. These word-level labels are gained through dependency parsing during the pre-processing step.
The branches separately integrate three co-dependent control factors: content, hallucination, and fluency.
Uncertainty-aware beam search (UABS)~\cite{xiao2021hallucination} is an extension to beam search to reduce hallucination. Considering that the hallucination probability is positively correlated with predictive uncertainty, this work adds a weighted penalty term in the beam search which is able to balance the predictive probability and uncertainty. \ziwei{This approach is task-agnostic and can also be applied to other tasks, such as image captioning.}

\ziwei{These various types of methods do not necessarily conflict and can collaborate to solve the hallucination problem in data-to-text generation. }

\subsection{Future Directions in Data-to-Text Generation}
Given the challenges brought by the discrepancy between structure data and natural text, and the low fault tolerance in the Data-to-Text Generation task, there are several potential directions worth exploring in terms of hallucination. 

Firstly, numbers contain information about scales and are common and crucial in the Data-to-Text task ~\cite{suadaa2021towards,zhang2020language}.
It is frequent to have errors in numbers, which results in hallucinations and infidelity. 
This is a serious problem for Data-to-Text generation, yet models rarely give special consideration to the numbers found in the table or text~\cite{thawani2021representing}. The current automatic metrics of hallucinations also do not specifically treat numbers. 
This indiscriminate treatment contradicts findings in cognitive neuroscience, where numbers are known to be represented differently from lexical words in a different part of the brain~\citep{gobel2004cognitive}.
Thus, considering or highlighting numbers when mitigating and assessing hallucinations is worth exploring. This requires the generative model to learn a better numerical presentation and capture scales, which will reduce the hallucinations caused by the misunderstanding of numbers. 

Moreover, for the logical data-to-text generation task, rather than surface-level generation, logical inference, calculation, and comparison are required, which is challenging and causes hallucinations more easily.   
Thus, reasoning (including numerical reasoning), which is usually combined with graph structures~\citep{chen2020logic2text} is another direction to improve the accuracy of entity relationships and alleviate hallucinations.

\section{Hallucinations in Neural Machine Translation}
\label{section:translation}
Neural Machine Translation (NMT) is the task of generating translation of the source language into the target language via inference, given parallel data samples for training. Compared to statistical machine translation (SMT) the output of NMT is usually quite fluent and of human-level quality, which creates the danger of misinforming users when there are hallucinations~\cite{ Martindale2019IdentifyingFI}. 

\subsection{Hallucinations Definition and Categories in NMT}
 The problem of hallucination was identified with the deployment of the first NMT models. Early work comparing SMT and NMT systems~\cite{Koehn2017}, without explicitly using the term “hallucination”, mentioned that NMT models tend to “sacrifice adequacy for the sake of fluency” especially when evaluated with out-of-domain test sets. Following further development of NMT, most of the relevant research papers agree that translated text is considered a hallucination when it is completely disconnected from the source ~\cite{lee2018hallucinations,muller2020domain}. 
  The categorization of hallucination in NMT is unlike that in any other NLG tasks, and uses various terms that are often overlapping. In order to maintain consistency with other NLG tasks, in this section we use the intrinsic and extrinsic hallucination categories applied to the NMT task by~\cite{zhou2021detecting}. After a formal definition, we will describe other identified types of hallucinations and hallucination categories mentioned in the relevant literature.
\paragraph{Intrinsic and Extrinsic Hallucinations}
Following the idea that hallucinations are outputs that are disconnected from the source,~\cite{zhou2021detecting} suggest categorizing the hallucinatory content based on the way the output is disconnected:
\begin{itemize}
    \item Intrinsic Hallucination: translations contain incorrect information compared to information present in the source. 
    In Table \ref{Table: categories_in_mt}, the example of such hallucination is “Jerry doesn't go”, since the original name in the source is “Mike” and the verb “to go” is not negated. 
    \item Extrinsic Hallucination: translations produce additional content without any regard to the source. In Table \ref{Table: categories_in_mt},  “happily” and “with his friend” are the two examples of the hallucinatory content since they are added without any apparent connection to the input.
\end{itemize}

\paragraph{Other Categories and Types of Hallucinations}
\citet{Raunak2021} propose an alternative categorization of hallucinations. They divide hallucinations into hallucinations under perturbations and natural hallucinations. Hallucinations under perturbation are those that can be observed if a model tested on the perturbed and unperturbed test set returns drastically different content. Their work on hallucinations under perturbation strictly follows the algorithm proposed by \citet{lee2018hallucinations}; see Section \ref{sec: entropy_measure} on the entropy measure. The second category, natural hallucinations, are created with a connection to the noise in the dataset and can be further divided into detached and oscillatory, where detached hallucinations mean that a target translation is semantically disconnected from a source input, and oscillatory hallucinations mean those that are decoupled from the source by manifesting a repeating n-gram. \citet{tu2016modeling} and \citet{kong2019neural} analyze this phenomenon under the name “over-translation”, that is, a repetitive appearance of words that were not in the source text. Conversely, under-translation is skipping the words that need to be translated~\cite{tu2016modeling}. Finally, abrupt jumps to the end of the sequence and outputs that remain mostly in the source language are also examples of hallucinatory content~\cite{lee2018hallucinations}.

\begin{table}[]
\resizebox{\textwidth}{!}{%
\begin{tabular}{@{}llll@{}}
\toprule
  \multicolumn{1}{c}{\textbf{Category}} &
  \multicolumn{1}{c}{\textbf{Source}} &
  \multicolumn{1}{c}{\textbf{Correct Translation}} &
  \multicolumn{1}{c}{\textbf{Hallucinatory Translation}} \\ \midrule
  Intrinsic &
  \begin{CJK*}{UTF8}{gbsn}  迈克周四去书店。\end{CJK*}   &
  \begin{tabularx}{0.4\textwidth}{X}Mike goes to the bookstore on Thursday.\end{tabularx} &
  \begin{tabularx}{0.4\textwidth}{X}Jerry doesn't go to the bookstore on Thursday.\end{tabularx} \\
  Extrinsic &
  \begin{CJK*}{UTF8}{gbsn}  迈克周四去书店。\end{CJK*}   &
  \begin{tabularx}{0.4\textwidth}{X}Mike goes to the bookstore on Thursday.\end{tabularx} &
  \begin{tabularx}{0.4\textwidth}{X}Mike happily goes to the bookstore on Thursday with his friend.\end{tabularx} \\\midrule
  Detached &
  \begin{tabularx}{0.4\textwidth}{X}Das kann man nur feststellen, wenn die kontrollen mit  einer großen  intensität  durchgeführt werden.\end{tabularx} &
  \begin{tabularx}{0.4\textwidth}{X}This can only be detected if controls undertaken are more rigorous.\end{tabularx} &
  \begin{tabularx}{0.4\textwidth}{X}Blood alone moves the wheel of history, i say to you and you will understand, it is a privilege to fight.\end{tabularx} \\
  Oscillatory &
  \begin{tabularx}{0.4\textwidth}{X}1995 das produktionsvolumen von 30 millionen pizzen wird erreicht.\end{tabularx}&
  \begin{tabular}[c]{@{}l@{}}1995 the production \\ reached 30 million pizzas.\end{tabular} &
  \begin{tabularx}{0.4\textwidth}{X}The US, for example, has been in the past two decades, but has been in the same position as the US, and has been in the United States.\end{tabularx} \\\bottomrule
\end{tabular}}
\caption{Categories and examples of hallucinations in MT by \citet{zhou2021detecting} and \citet{Raunak2021}}
\label{Table: categories_in_mt}

\end{table}


\subsection{Hallucination Metrics in NMT}

The definition of hallucinations in machine translation (MT) tends to be qualitative and subjective, and thus researchers often identify hallucinated content manually. Most detrimentally, the appearance of hallucinations is found not to affect the BLEU score of the translated text~\cite{tian2020sticking,zhou2021detecting}. There are, nevertheless, several notable efforts to automatize and quantify the search for hallucinations using statistical methods.

\subsubsection{Statistical Metrics}

\citet{Martindale2019IdentifyingFI} propose identifying sentence adequacy using the bag-of-vectors sentence similarity (BVSS) metric. This metric indicates that the information is lost because the reference contains more information than the MT output, or the MT output contains more information than the reference. 

\subsubsection{Model-Based Metrics}
\paragraph{Auxiliary Decoder}\label{sec: auxiliary decoder}
“Faithfulness” refers to the amount of source meaning that is faithfully expressed in the translation, and it is used interchangeably with the term “adequacy”~\cite{tu2017neural,feng2020modeling}. \citet{feng2020modeling} propose adding another “evaluation decoder” apart from the standard translation decoder.  In their work, faithfulness is based on word-by-word translation probabilities, and is calculated in the evaluation module along with translation fluency. The loss returned by the evaluation module helps to adjust the probability returned by the translation module.

\paragraph{Entropy Measure}\label{sec: entropy_measure}
In scenarios where the ground truth of a translation is not available, an entropy measure of the average attention distribution can be used to detect hallucinations. \citet{tu2016modeling} and \citet{garg2019jointly} show that hallucinations are visible in attention matrices. When the model outputs correct translation, the attention mechanism attends to the entire input sequence throughout decoding. However, it tends to concentrate on one point when the model outputs hallucinatory content. 
The entropy is calculated on the average attention weights when the model does or does not produce hallucinations during testing. 
For comparison, a clean test set is used along with the purposefully perturbed one, which is created to incite hallucinations (test sets featuring multiple repetitions). 
The mean entropy returned by hallucinatory models diverges from the mean of the models that do not produce hallucinations spontaneously~\cite{lee2018hallucinations}.


\paragraph{Token Level Hallucination Detection}
\label{paragraph: token_level}

\citet{zhou2021detecting} propose a method for detecting hallucinated tokens within a sentence, making the search more fine-grained. They use a synthetic dataset that is created by adding noise to the source data, more specifically it is generated by a language model with certain tokens of correct translations masked. Tokens in synthetic data are labeled as hallucinated (1) or not (0). Then the authors compute the hallucination prediction loss between binary labels and the tokens from the hallucinated sentence.
This work further employs a word alignment-based method and overlap-based method as baselines for hallucination.

\paragraph{Similarity-based Mathods}
\citet{zhou2021detecting} use an unsupervised model that extracts alignments from similarity matrices of word embeddings~\cite{sabet2020simalign}, and then predicts the target token as hallucinated if it is not aligned to the source.
\citet{parthasarathi2021want} propose calculating faithfulness by computing similarity scores between perturbed source sentence and target sentence after applying the same perturbation.

\paragraph{Overlap-based Mathods}
\citet{zhou2021detecting} predict that the target token is hallucinated if it does not appear in the source. Since the target and source are two different languages, the authors use the density matching method for bilingual synonyms from \citet{Zhou2019DensityMF}.
\citet{kong2019neural} suggest the Coverage Difference Ratio (CDR) as the metric to evaluate adequacy, which is especially successful in finding cases of under-translation. It is estimated by comparing source words covered by generated translation with human translations. 

The overlap-based methods for detecting hallucinations are heuristics based on the assumption that all translated words should appear in the source. However, this is not always the case, e.g., when paraphrasing or using synonyms. Using word embeddings as similarity-based methods helps avoid such simplifications and allows more diverse, synonymous translations. 

\paragraph{Approximate Natural Hallucination Detection} \citet{Raunak2021} propose Approximate Natural Hallucination (ANH) detection based on the fact that hallucinations often occur as oscillations (repeating n-grams) and the lower unique bigram count indicates a higher appearance of oscillatory hallucinations. Furthermore, the ANH detection method searches for repeated targets in the translation output. Their method finds translation above a certain n-gram threshold and searches for repeated targets in the output translation, following the assumption that if hallucinations are often incited by aligning unique sources to the same target, then repeating targets will also appear during the inference~\cite{tu2016modeling}.

\subsection{Hallucination Mitigation Methods in NMT}

Hallucinations in MT are hard to discover for a person who is not fluent in the target language, and thus they can lead to many possible errors, or even dangers. Out of all the natural language generation tasks, NMT engines such as Google in the English-speaking internet and Baidu in the Sinosphere are probably the most widely accessible to netizens. Consequently, there is a big interest in improving NMT´s performance, also by mitigating hallucinations. This subsection compiles  methods of mitigating hallucinations in NMT.

\subsubsection{Data-Related} Data augmentation appears to be one of the most common methods for removing hallucination. \citet{lee2018hallucinations} and \citet{Raunak2021} suggest addition of perturbed sentences. Furthermore, perturbation, where the insertions of most common tokens are placed at the beginning of the sentence, seems to be the most successful in hallucination mitigation. A disadvantage of this method is the need to understand different types of hallucinations produced by the model in order to apply a correct augmentation method.
Corpus filtering is a method of mitigating hallucinations caused by the noise in the dataset by removing the repetitive and mismatching source and target sequences~\cite{Raunak2021}. 
\rita{\citet{junczysdowmunt2019dual} implements a cross-entropy data filtering method for bilingual data, which uses cross-entropy scores calculated for noisy pairs according to two translation models trained on the clean data. The scores that suggest dissagreament between sentence pairs from two models are subsequently penalized.}

While~\cite{Raunak2021,lee2018hallucinations} and~\cite{junczysdowmunt2019dual} define noise as mismatched source and target sentences, ~\cite{briakou2021} analyzes the influence of fine-grained semantic divergences on NMT outputs. The authors consequently propose a mitigation method for fine-grained divergences based on semantic factors. The tags are applied to each source and target sentence to inform about the position of divergent tokens. Factorizing divergence not only helps to mitigate hallucinations, but improves the overall performance of the NMT. This shows that tagging small semantic divergences can provide useful information for the network during training.

\subsubsection{Modeling and Inference} 
Overexposure bias is a common problem in NMT, amplified by the teacher-forcing technique used in sequence-to-sequence models. The models are trained on the ground truth, but during inference, they attend to the past predictions, which can be incorrect~\cite{ranzato2016sequence, kong2019neural}.
To mitigate this problem, \citet{wang2020exposure} propose substituting MLE as a training objective with minimum risk training (MRT)~\cite{och2003minimum}. 
Scheduled sampling is a classic method of mitigating overexposure bias first proposed by~\cite{bengio2015scheduled}. Based on that method,~\cite{Goyal2017} create a differentiable approximation to greedy decoding that shows a good performance in the NMT task.~\cite{Xu2019} propose further improvement of the scheduled sampling algorithm for NMT by optimizing the probability of source and target word alignments. This improvement helps to address the issue flexibility in word order between a source and target language when performing scheduled sampling.

\citet{zhou2021detecting} propose a method of improving self-training of NMT based on hallucination detection. They create hallucination labels (see Section \ref{paragraph: token_level}), and then discard losses of tokens predicted as hallucinations, which is known as token loss truncation. This is similar to the method proposed by \citet{Kang2020}, the latter for full sentences in the summarization task. Furthermore, instead of adjusting losses, \citet{zhou2021detecting} mask the hidden states of the discarded losses in the decoder in a procedure called decoder HS masking. Experimental results show both a translation quality improvement in terms of BLEU and also a large reduction in hallucination. The token loss truncation method shows good results in the low-resource languages scenario.

Another method to mitigate the impact of noisy datasets is tilted empirical risk minimization (TERM), a training objective proposed by \citet{Li2021_TERM}.
\citep{lee2018hallucinations} mentions that techniques such as dropout, L2E regularization, and clipping tend to decrease the number of hallucinations. Lastly, several authors propose methods of improving phrase alignment that are helpful both in increasing translation accuracy and identifying content that did not appear in the source translation~\cite{weng2020towards, zhang2021neural, garg2019jointly}. 

\subsection{Future Directions in NMT}
The future work on hallucinations in NMT is to define hallucinations in a quantifiable manner; i.e., to specify a cut-off value between translation error and hallucinated content using a particular metric. 
\citet{Martindale2019IdentifyingFI} propose a threshold between fluency and adequacy which is the closest to this ideal. They, however, do not concentrate on hallucinated content as such, and thus fluent but inadequate sentences may not always indicate hallucinations but also other types of translation errors.
\citet{balakrishnan-etal-2019-constrained} mention constrained decoding as a method to mitigate hallucinations in dialogue systems, but it could also be applied in NMT.~\cite{Hokamp2017,Post2018,Song2019,Dinu2019,Susanto2020,Xu2021} and~\cite{Xu2020} use constrained decoding to incorporate specific terminology into MT, but the above methods can be repurposed to mitigate hallucinations. 

Another direction for future work on hallucinations is improving existing methods of searching for hallucinatory content, such as the algorithms proposed by \citet{lee2018hallucinations, feng2020modeling} and \citet{Raunak2021}, that are computationally expensive~\cite{Raunak2021} or require the creation of an additional perturbed test-set~\cite{lee2018hallucinations}. 
Similarly, for mitigation of lack of faithfulness and fluency, the method proposed by \citet{feng2020modeling} requires the creation of a one-to-many architecture (one encoder and two decoders), which is also computationally expensive. 
Future directions would therefore include simplification of existing hallucination evaluation methods, applying them to different architectures like CNNs and transformers, and possibly conducting research on finding simpler hallucination search methods.

\section{Hallucination in Vision-Language Generation}
\label{section:VL}

With the vast advancement of the Transformer architecture~\citep{vaswani2017attention,dosovitskiy2021an} in both CV and NLP, there is a trend to pre-train large-scale unified vision-language (VL) models~\citep{wang2022ofa,dai-etal-2022-enabling,li2022blip,Wang2022SimVLMSV,alayrac2022flamingo,Wang2022ImageAA} to perform vision grounded text generation tasks, such as image captioning and visual question answering (VQA). Generally, there are two common schemas for vision-language pre-training: 1) pre-train from scratch with a massive amount of image-text pairs as well as optionally a large text-only corpus; or 2) initialize model parameters from a large pre-trained LM and then adapt it to the VL domain with adequate image-text pairs. Either way, the learned vision and language representations are aligned in the same multimodal space and the resulting model can be seen as a LM that understands visual information. Therefore, the hallucination problem is also observed in VL models due to similar reasons as found in NLG. 

In the VL domain, the research on hallucination is still in its very early stage and how to measure and mitigate hallucination is an open question. In this section, we first review the hallucination in image captioning as it is the only VL task that has corresponding previous research works. Then, we introduce hallucination phenomena found in other VL tasks. Finally, we discuss potential future research directions on this problem.


\subsection{Object Hallucination in Image Captioning}

\paragraph{Definition.} Object hallucination is defined as models generating captions that contain non-existent or inaccurate objects from the input image. Following tasks in NLG, we also categorize object hallucination into intrinsic and extrinsic ones:

\begin{figure}[ht]
    \centering
    \includegraphics[width=0.65\linewidth]{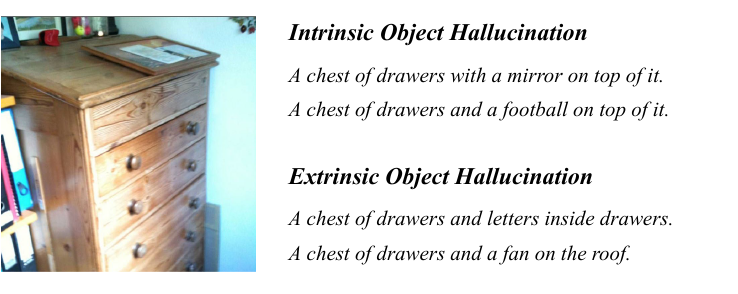}
    \caption{Examples of intrinsic and extrinsic object hallucination in image captioning.}
    \label{fig:captioning_examples}
\end{figure}

\begin{itemize}
    \item Intrinsic Object Hallucination: captions contain incorrect or definitely non-existent objects given the input image. For example in Figure \ref{fig:captioning_examples}, there is no \textit{``mirror''} or \textit{``football''} on top of the chest in the given image.
    \item Extrinsic Object Hallucination: captions contain objects cannot be verified their existence from the input image. For example in Figure \ref{fig:captioning_examples}, we cannot verify whether there are \textit{``letters''} in the drawer or a \textit{``fan''} on the roof.
\end{itemize}

\paragraph{Metrics} To automatically measure object hallucination, \citet{rohrbach2018object} propose the CHAIR (Caption Hallucination Assessment with Image Relevance) metric, which calculates what proportion of object words generated are actually in the image according to the ground truth captions. Specifically, there are two variants of it, which are CHAIR$_i$ and CHAIR$_s$ defined as follows,

\begin{align*}
    \textrm{CHAIR}_i = \frac{\textrm{\# \{hallucinated objects\}}}{\textrm{\# \{all objects in ground truth\}}}, 
    \textrm{CHAIR}_s = \frac{\textrm{\# \{hallucinated captions\}}}{\textrm{\# \{all captions\}}}.
\end{align*}

\noindent CHAIR$_i$ measures per-instance object hallucination, i.e. what fraction of object instances in each generated caption are hallucinated. CHAIR$_s$ measures per-sentence object hallucination, i.e. what fraction of generated captions include at least one hallucinated object. For example, to calculate CHAIR scores for the MSCOCO dataset~\citep{Lin2014MicrosoftCC}, \citet{rohrbach2018object} apply the 80-object list used in the MSCOCO segmentation challenge \citet{Lu2018Neural} and find exact matches of object words or phrases in captions.


\paragraph{Mitigation} As a research problem that is in its early stage, there are currently a limited number of approaches proposed to mitigate object hallucination in image captioning. \citet{biten2022let} hypothesize that the main cause of object hallucination is the systematic co-occurence of particular object categories in input images. They propose three simple yet effective ways of data augmentation to make the co-occurence statistics matrix more uniform to mitigate object hallucination. Results show that their introduced method can reduce object hallucination without changing model architectures.  From another perspective, \citet{xiao2021hallucination} propose an uncertainty-aware beam search method for decoding and exhibit that reducing uncertainty can lead to less hallucination. Specifically, a weighted penalty term is added to the beam search objective to balance between log probability and predictive uncertainty of the selected word candidates. More recently, \citet{Dai2022PlausibleMN} analyze object hallucination in VL pre-training and propose a novel pre-training objective named object masked language modeling to alleviate this problem.

\subsection{Hallucination in Other VL Tasks} \label{sec:hall_VL_other}
\begin{figure}[h]
    \centering
    \includegraphics[width=0.9\linewidth]{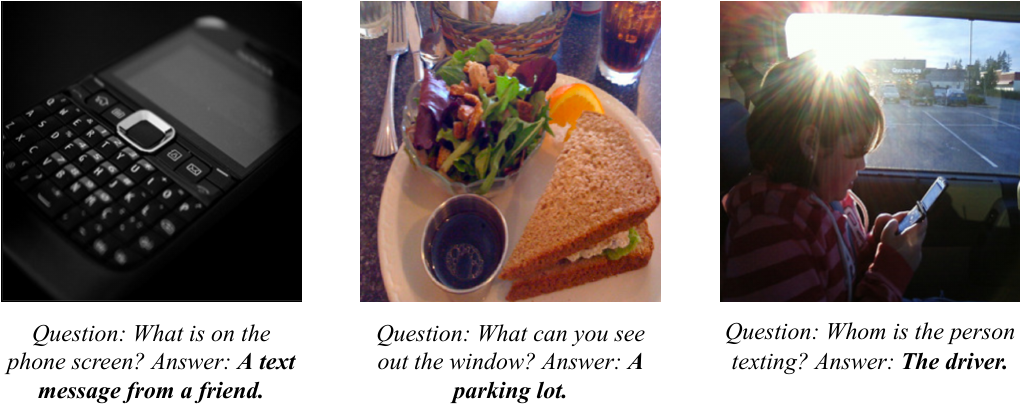}
    \caption{Examples of hallucination in visual question answering (taken from \citep{alayrac2022flamingo}). The bold text is the output generated by the model and the part before it is the input prompt.}
    \label{fig:vqa_examples}
\end{figure}

In addition to image captioning, hallucination has also been observed in other VL tasks and raised as an open research question. For example, in open-ended visual question answering, Figure~\ref{fig:vqa_examples} (left and right) shows that the model could generate seem likely answers when we only see the text, however wrong when given the image. Moreover, Figure~\ref{fig:vqa_examples} (middle) indicates that hallucination can also be triggered by adversarially prompting an unanswerable question. The model will imagine an unsupported answer that commonly matches the given visual scene. 

\subsection{Future Directions in VL}
For future research on the hallucination problem in VL, we summarize three promising directions. 
Firstly, hallucination detection and mitigation in VL is still in the early stage. There is a lack of empirical and theoretical analyses in many tasks, such as visual storytelling, visual commonsense reasoning, video captioning, etc. Secondly, more effective evaluation metrics are needed. For example, although CHAIR can automatically evaluate the degree of object hallucination in image captioning, it requires a pre-defined list of object categories, which does not generalize well. Furthermore, currently there is no automatic metric for the hallucination types discussed in Section~\ref{sec:hall_VL_other}. Therefore, we cannot perform quantitative evaluations for them. Thirdly, we believe how to perform controlled generation~\citep{rebuffel2022controlling,Dai2022PlausibleMN} with visual grounding is a promising direction to mitigate hallucination in VL.

\section{Hallucination in Large Language Models}
\label{section: LLM}
Scaling up models and data sizes for language models has shown great empirical success. The resulting Large Language Models (LLMs) not only have achieved significant performance improvements upon previous PLMs across a diverse array of NLP tasks, but have also demonstrated many emerging abilities and strong steerability after instruction tuning and Reinforcement Learning from Human Feedback (RLHF)~\cite{zhao2023survey, wei2022emergent}. However, they still exhibit the hallucination problem. Even worse, since LLMs generate highly fluent and convincing responses, their hallucinations become more difficult to identify, and more likely to have harmful consequences. The launch of ChatGPT as an LLM with a conversational user interface led to the popularity of LLMs and their wide range of real-world applications. The research on the identification and mitigation of hallucination in LLMs has intensified. LLMs are usually open-ended general-purpose systems, which differ much from task-specific models, and their architectural designs, data coverage, training methodologies, and model behaviors are also different from PLMs mentioned in previous sections. Therefore, in the following sections, we specifically discuss the hallucination problem in LLMs, by covering more recent works, introducing novel strategies for measuring and mitigating hallucination, as well as listing unsolved questions and future directions.~\footnote{This section was updated in Jan 2024.}

\subsection{Hallucination Definition in LLMs}
In the realm of LLMs, hallucination takes on a broader definition than before because of the vastness of training data, the breadth of knowledge base, and multitasking capability. 
Unlike earlier NLG models that depend heavily on the input source, LLMs leverage their vast internal parametric knowledge.
Consequently, hallucination in LLMs not only signifies deviations from the source input but also extends to deviations from training data, marking it more oriented towards the \textbf{Extrinsic} type. 
In other words, the reference discussed in Section~\ref{subsec:terminology} includes both the input source and the training data. 

Existing works often assume that training data is consistently factual and treat it as a stand-in for ``fact'' or ``truth''.
This assumption allows for a practical simplification, with the degree of hallucination reflecting the model's ability to accurately and faithfully comprehend and represent the world.
However, this assumption is not always valid, given that training datasets may contain non-factual and contradictory information.
Thus, the concept of hallucination should be distinguished from that of factuality.


\subsection{Hallucination Metrics for LLMs }
Hallucination metrics for LLMs are designed to either measure the level of hallucination in LLM-generated content, or quantify the hallucination risk of individual LLMs. Most of the related work focuses on measuring extrinsic hallucinations in the open domain. Depending on whether a metric requires golden references to identify hallucinations, we divide existing hallucination metrics for LLMs into reference-dependent and reference-free ones. In reference-dependent metrics, generations of LLM are analyzed based on references sourced from verified knowledge bases such as Wikipedia, news articles, books, and knowledge graphs. For reference-free metrics, hallucinations are identified by token probability-based uncertainty measures or self-consistency checks.

\subsubsection{Reference-dependent Metrics}

    Reference-dependent metrics are often formatted as ``hallucination benchmarks'' that measure the hallucination level not risks of different LLMs. Most benchmarks utilize multiple choice QA or fill-in-the-blank QA to assess model hallucination, while some efforts further consider the ratio of partially incorrect answers~\cite{Sun2023Head} or instances where the model admits that it cannot answer~\cite{Lin2022TruthfulQA}. Some benchmarks adopt the open domain and open-ended QA setting, where exact-match-based evaluations generally fail to recognize semantically equivalent answers~\cite{Kamalloo2023Evaluating}. Free-form responses from LLMs can be analyzed based on atomized decomposition (FactualityPrompt~\cite{lee2022factuality} and FActScore~\cite{Min2023FActScore}), LLM-based evaluation (self-judgment~\cite{Ren2023Investigating} or GPT-4-based judgment~\cite{Adlakha2023Evaluating}), or by measuring the discrepancy of the vanilla response and the response giving the knowledge for reference~\cite{Yu2023KoLA,Huo2023Retrieving,Pezeshkpour2023Measuring}.

    \paragraph{Human-annotated Hallucination Benchmarks} 
    \label{SEC:Human-annotated Hallucination Benchmarks} 
    Collecting human-verified QA samples to test the factual knowledge of LLMs is the most straightforward method. Since the training corpora of LLMs are extremely large, high frequency count knowledge and basic commonsense can be easily memorized. As a result, examining low frequency, long-tailed knowledge becomes a primary focus of these hallucination benchmarks. Testing samples for long-tail knowledge can be selected based on frequency of appearance (\textit{e.g.,} in pretraining corpora~\cite{Kandpal2023Large} or Wikitext~\cite{Du2023Quantifying}), popularity measurements based on website traffic data (\textit{e.g.,} PopQA~\cite{Mallen2023When}, Head-to-Tail~\cite{Sun2023Head}), recency (RealTimeQA~\cite{Kasai2022RealTime}), or from specific domains (\textit{e.g.,} ExpertQA~\cite{Malaviya2023ExpertQA} and Med-HALT~\cite{Umapathi2023Med}). In addition to long-tail knowledge, another way to build challenging benchmarks is to gather questions that are more likely to lead to hallucinations -- a methodology similar to adversarial prompting~\cite{Kumar2023Certifying,zhu2023promptbench} and red teaming~\cite{Ganguli2022Red}. The most representative example is TruthfulQA~\cite{Lin2022TruthfulQA}, a benchmark containing challenging questions that some humans would answer falsely due to false beliefs or misconceptions (\textit{e.g.,} question \textit{``If it's cold outside, what does that tell us about global warming?''} leads to GPT-3 answers \textit{``It tells us that global warming is a hoax''}). 
    Similarly, the DecodingTrust~\cite{Wang2023DecodingTrust} evaluated the trustworthiness of GPT models in terms of adversarial robustness, OOD knowledge, vulnerability of jail-breaking, misleading instructions, etc.
    Known-Unknown~\cite{Amayuelas2023Knowledge, azaria2023internal} and SeflAware~\cite{Yin2023Do} evaluated LLMs' awareness of uncertainty for question without definitive answer (\textit{e.g.,} \textit{``If the Universe started at the Big Bang, what existed before then?''})

    \paragraph{Automated Hallucination Benchmarking} Manual question collection and answer annotation are labor intensive, hard to scale, and hard to adapt to new domains. To address this issue, annotation-free benchmarking methods have been proposed, which can be further divided into 1) automated benchmark generation and 2) automated evidence retrieval. For the first type of methods~\cite{Muhlgay2023Generating,Chen2023Benchmarking}, factual but unstructured corpora (for example, Wikipedia and reliable news articles) are transformed into a unified QA format based on information extraction and answer candidate generation. The second type of methods automatically retrieves evidence related to LLM generations from the Internet, similar to automated fact checking~\cite{Thorne2018Automated,Guo2022Survey}. They typically adopt a chained pipeline, where search queries are first generated (by an LLM) based on the content that needs to be verified, and then a model determines whether it contains hallucinations according to the retrieved evidence~\cite{Gao2023RARR}). This pipeline can be further enhanced by upgrading the verification model from an NLI model~\cite{Gao2023RARR} to prompted/fine-tuned LLMs (as in AttributionScore~\cite{Yue2023Automatic} and~\cite{Varshney2023Stitch}), or by adding additional steps to the pipeline, such as tool interaction (as in FactTool~\cite{Chern2023FacTool} and CRITIC~\cite{Gou2023CRITIC}), claim decomposition (Self-Checker~\cite{Li2023Self}), additional second stage fine-grained retrieval and claim-focused summarization (as in~\cite{Chen2023Complex}), or chain-of-verification~\cite{Dhuliawala2023Chain}.
    
\subsubsection{Reference-free Metrics}

    Reference-free metrics aim to assess the hallucination risk of LLMs without relying on external references or ground truth. Although not as reliable as reference-based ones, these metrics are particularly useful when golden references are unavailable, expensive to obtain, or when hallucination identification needs to be performed in real-time. Reference-free metrics generally fall into two categories: uncertainty-based and consistency-based hallucination detection. Uncertainty-based methods rely on the token probabilities assigned by the LLM during generation, whereas consistency-based methods evaluate the coherence of multiple completions generated by the LLM. There are also some benchmarks to measure the effectiveness of these reference-free hallucination metrics, which are also termed ``hallucination detection benchmarks'', such as HaluEval~\cite{Li2023HaluEval} and HaDes~\cite{Liu2022Token} These benchmarks serve as meta-metrics of hallucination metrics, offering valuable resources for developing more effective methods to identify hallucinations.

    \paragraph{Uncertainty-based Hallucination Detection} This type of method assumes that LLMs assign high probabilities to tokens that they are confident of, and low probabilities to uncertain tokens, which usually contain hallucinated information. It has a close relationship with classical uncertainty estimation methods~\cite{Abdar2020Review,Mena2022Survey}, since the generation of each token can be viewed as a vocabulary size classification problem. The difference here is that autoregressive text generation of LLM is a chained classification process, therefore individual token (subword) probabilities need to be aggregated so that they can reflect word-/sentence-/passage-level uncertainty. Such aggregation can be done by average-/max-/min-pooling on token probabilities, calculating the normalized product of the token probabilities, or calculating the maximum/averaged entropy~\cite{Manakul2023SelfCheckGPT,Varshney2023Stitcha,Huang2023Look,Kadavath2022Language}. These methods can be further extended by adding prompts like \textit{``Generate factually consistent summary for the following text: {<source-text> <generated-text>}''}~\cite{Fu2023GPTScore} for fine-grained control (factuality evaluation on summarization), or \textit{``{<question> <generated-answer>} The answer is true.''}~\cite{Kadavath2022Language} for self-evaluation of QA samples.

    \paragraph{Consistency-based Hallucination Detection} is the more reliable and common approach. In practical applications, many LLMs only offer an API while internal logits remain inaccessible (\textit{i.e.,} the ``black-box setting''), thus making uncertainty-based methods inapplicable. Efforts have been made to create surrogate approximations of token probability. Specifically, LLMs first generate multiple completions with stochastic sampling given a fixed context, and then the consistency of these completions is used to reflect the uncertainty. Consistency can be measured by the BLUE-based variation ratio~\cite{Huang2023Look}, BERTScore~\cite{Manakul2023SelfCheckGPT,Raj2022Measuring}, n-gram approximation~\cite{Manakul2023SelfCheckGPT}, NER-based overlap ratio~\cite{Raj2022Measuring}, NLI model~\cite{Raj2022Measuring,Elaraby2023Halo}, or LLM-based judgment~\cite{Muendler2023Self}. 
    

\subsection{Hallucination Mitigation in LLMs}
\subsubsection{Data-Related Methods} 
\label{sec:llm:method:data}
\paragraph{Data for Pretraining} 
Recent studies~\cite{zhou2024lima} show that the knowledge within LLMs is almost entirely acquired during pretraining. Thus, a strong emphasis should be placed on ensuring the quality of the pretraining data. This can be performed through the collection of pretraining data from credible sources~\cite{LLaMA2023} and the minimization of defective and noisy data such as those that are unreliable or unverifiable~\cite{SirenSong2023}. A representative example of such data-driven methods can be found in the development of Llama-2~\cite{LLAMAv2}, where the most factual sources within the pretraining data were up-sampled in order to reduce hallucinations. Similarly, Falcon LLM~\cite{penedo2023falcon} also demonstrated that their data refinement methods including rigorously filtering and deduplicating can significantly boost LLM performance.  Due to the extensive pretraining datasets, some approaches have been developed to overcome the impracticalities that arise from manually selecting them. For instance, for the pretraining of GPT-3~\cite{brown2020language}, a classifier-based automatic filtering method was applied to remove low-quality documents. However, as pretraining data is being continuously scaled up, it is becoming increasingly costly for the research community to even access the entire pretraining dataset. As an alternative, small LLMs like phi-1.5~\cite{TextAAYN_tech_report}, which are trained on strictly controlled ``textbooks-style'' small-scale corpus, provide a valuable opportunity for doing rigors ablations to investigate the relationship between pretraining data and model hallucination.
    
    \paragraph{Data for Instruction-tuning} During the fine-tuning stage, data refinement can also be used to dampen hallucinations. As the size of fine-tuning data is significantly smaller, manual approaches can also be practically used~\cite{SirenSong2023}. For example, the authors of LIMA collected a diverse dataset of 1,000 carefully curated prompts and responses which are aligned with each other~\cite{zhou2024lima}. Conversely, \citet{AlpaGasus2023} point out that widely used fine-tuning datasets e.g. Alpaca’s 52k data contain low-quality instances with incorrect or irrelevant responses which are detrimental to fine-tuning, used an automated approach leveraging strong LLMs to identify and discard such low-quality data. Alongside, the authors of InstructMining~\cite{InstructMining2023}, proposed linear rules for selecting high-quality instruction fine-tuning data and avoided the need for human or machine annotations. 

    \paragraph{Data for Reward Model Training} Training reliable reward models such that desirable and undesirable outputs are distinguished is an effective method for mitigating hallucination. These models can subsequently be incorporated into reinforcement learning pipelines. Since the effectiveness of the system is constrained by the performance of the reward model itself, emphasis has been placed on determining the most effective training methods~\cite{VerifyStepByStep2023}. Thus, OpenAI released a dataset, namely, PRM800K, which resulted in a SOTA reward model~\cite{VerifyStepByStep2023}. In the development of GPT-4~\cite{GPT4_2023}, two different approaches were implemented. Firstly, in tackling open-domain hallucinations, OpenAI collected real-world ChatGPT data that were flagged as non-factual by users and used it together with additional labeled comparison data to train their reward models. Secondly, for closed-domain hallucinations, OpenAI utilized GPT-4 itself to generate and subsequently utilize synthetic data for reward model training.

\paragraph{Retrieval Augmented Generation (RAG)}
RAG is a technique that leverages external sources to enhance the reliability of generative models~\cite{IBM_RAG2023, zhang2023mitigating}, which is a popular method used to mitigate hallucination~\cite{RAGAS2023}. For example, Gemini incorporates the search engine to provide external references during generation. 
However, RAG faces several challenges~\cite{Chen2023Benchmarking}: 
In the retrieval stage, the model is required to discern the noise, irrelevant or fake information. 
In the augmentation stage, integrating heterogeneous, independent, and even conflicting information presents challenges. 
Additionally, the model needs to reject generation when insufficient information is given.
Recent works make efforts to address these challenges.
For example, MixAlign~\citet{zhang2023mitigating} enhances the alignment between user questions and stored knowledge to improve retrieval and reduce hallucination.

\subsubsection{Modelling and Inference Methods}
\paragraph{Safety Fine-Tuning and Reinforcement Learning with Human Feedback (RLHF)}
Safety fine-tuning is a prevalent method to mitigate risks associated with LLMs, such as hallucinations. A prominent technique in this domain is Reinforcement Learning with Human Feedback (RLHF).
The application of RLHF in the fine-tuning process of GPT-3 leads to the development of InstructGPT, which significantly reduces the hallucination rate from 41\% to 21\% in closed domain tasks~\cite{InstructGPT2022}.
Similarly, Llama2~\cite{LLAMAv2} demonstrates that post-RLHF, the model consistently generated factual responses to prompts on factual information. The study also highlighted the use of RLHF in enhancing the overall safety of LLMs, including a reduction in hallucinations~\cite{LLAMAv2}.
Despite the advantages, Safety Fine-Tuning and RLHF can incur `alignment tax' and `catastrophic forgetting' where the models lose diverse previously acquired abilities after alignment~\citep{lin2023speciality}. To counterbalance these issues, \citet{lin2023speciality} proposes Adaptive Model Averaging (AMA) optimizing the averaging ratios of different model layers to maximize alignment reward. 

\paragraph{Model Editing}
Editing Models such as parameter adapting can also serve to reduce hallucination.
Given that some parameters are more responsible for causing hallucinations, EWR (Elastic Weight Removal)~\cite{EWR2023} weighs their individual importance via Fisher Information matrix and performs parameter interpolation to remove the undesired behaviors.

\paragraph{Decoding Methods}
Without editing architecture or additional fine-tuning, novel decoding strategies are proposed to reduce hallucinations.
Inspired by the finding that factual knowledge is localized to particular layers, DoLa (Decoding by Contrasting Layers)~\cite{DOLA2023} utilizes the difference in logits from the later layers versus earlier layers to amplify the factual knowledge probability. 
Similarly, CAD (Context-Aware Decoding)~\cite{CAD2023} exploits the difference in logits with and without context.
In addition,~\citet{li2024inference} observe the gap between surface generation and internal knowledge and propose ITI (Inference-Time Intervention) to reduce this gap. It first identifies attention heads with high linear probing accuracy for truthfulness and then shifts activations along these truth-correlated directions during inference.

\paragraph{Chain-of-Thought and Variants}
Chain-of-Thought and its variants encourage LLMs to reason prior to arriving at a final reply, which improves performance and reduces hallucinations by leveraging the model's internal knowledge~\cite{kojima2023, Dhuliawala2023Chain, lei2023chain, ji2023towards, shinn2023reflexion, wang2023unleashing}. 
For example, Chain-of-Verification (COVE)~\cite{Dhuliawala2023Chain} consists of four steps where an LLM first drafts an initial response, plans verification questions to fact check it, answers the questions independently to ensure the answers are not biased by other responses, and finally generates the verified response. 
Chain of Natural Language Inference (CoNLI)~\cite{lei2023chain} detect hallucinations in the initial response by prompting the LLM to conduct sentence/entity-level NLI, and then refine the response according to the detection result.
\citet{ji2023towards} propose iterative self-reflection loops that involve generating relevant background knowledge for a given question followed by a factuality evaluation and self-correction. During answering, a similar generation-score-refine strategy is used for its entailment with the question.
Solo Performance Prompting (SPP)~\cite{wang2023unleashing} prompts a single LLM to engage in multi-turn self-collaboration with multiple personas, such that their individual strengths and knowledge are combined to boost the problem-solving and performance in tasks. 

\paragraph{Post-processing}
Petite Unsupervised Research and Revision (PURR)~\cite{PURR2023} fine-tune an editor to denoise the corruptions and use the editor to correct the hallucinations in the initial outputs.
Other works like REFEED~\cite{REFEED2023} leverage retrieved documents for post-correction.
Similarly, CRITIC~\cite{Gou2023CRITIC} uses external tools such as search engines and code interpreters to verify and post-correct the initial output. 

\paragraph{Ensemble}
Ensembling multiple models to mitigate hallucination is another way. 
For example, in~\cite{duimproving}, multiple LLMs individually produce responses and then jointly debate their responses and reasoning, ultimately reaching a single common answer. This approach has been demonstrated to improve the factuality of generated content.


\subsection{Future Directions of Hallucination Mitigation in LLMs}
 
\subsubsection{Hallucination in Large Multimodal Models}
    Large multimodal models (LMMs) such as GPT-4V can generate rich and detailed responses for visual inputs, therefore being inherently riskier compared to previous VLMs for single-sentence captioning and short-answer VQA. Furthermore,  images convey different levels of information encompassing object existence, attributes (color, shape, etc), spatial relationships, and sometimes high-level emotional responses they elicit (peaceful, beautiful, etc.), which makes mitigating hallucination in LMM even more challenging. The exploration of this area is still in its early stage, with a primary focus on object existence hallucination~\cite{DBLP:conf/emnlp/LiDZWZW23}. However, it is imperative for LMMs to analyze images well beyond that surface level, otherwise, the use of LMM would be redundant and a simple object detector would suffice. The most crucial factor in this problem is the availability and quality of data and supervision. Language-supervised representations (\textit{e.g.,} CLIP) have demonstrated inherent shortcomings in recognizing certain types of visual patterns, such as object orientation, quantity, and viewpoint~\cite{tong2024eyes}. Even with perfect representation, imperfect vision-language alignment can result in a significant gap between the visual recognition abilities of the LMM and its visual backbone~\cite{zhai2023investigating}. The cost of annotation during the mitigation of multimodal hallucination should also be considered. While there are works on the visual extension of faithfulness-oriented RLHF~\cite{DBLP:journals/corr/abs-2312-00849}, these essentially involve labeling more data, which is expensive and difficult to scale. The versatility of the visual domains also makes it challenging to guarantee robustness and generalization. Large-scale self-supervision may be the path forward, as supervised learning has proven to be insufficient for both the CV and NLP in the past decade.

\subsubsection{Hallucination in Long-tail and Low-resource Domains}
   Hallucinations in LLMs in long-tail and low-resource domains still remain a significant challenge. This has become a new research focus -- for example, many recent hallucination benchmarks are proposed to measure LLM hallucination in various long-tail and low-resource domains (see Section~\ref{SEC:Human-annotated Hallucination Benchmarks}). However, our scope should not be confined to only narrow professional areas (\textit{e.g.,} medical, legal, finance), but should also consider broader scenarios that are also long-tailed and low-resource. For example, hallucination in low-resource languages is a less explored area, as hallucination mitigation methods applied to rich-resource languages do not necessarily generalize to low-resource ones. Similarly, the temporal axis also impacts the frequency of world knowledge appearance, where more recent data could also be considered as long-tailed and low-resource. Furthermore, beyond textual data, multi-modal scenarios are inherently low-resource for text-only pretrained LLMs. In light of the above, instead of developing domain-specific methods for hallucination mitigation, we need to design a unified framework and generalizable strategies for addressing hallucination in various imbalanced scenarios.

\subsubsection{Estimating the Knowledge Boundary and Expressing the Uncertainty}
    Despite the vast size and continuous scaling of pretraining datasets for LLMs, it is impossible for them to encompass all the world's knowledge. Consequently, teaching models to express their inability to answer specific questions and honestly admit ``I don't know'' is crucial, and it requires LLMs to accurately model their knowledge boundaries. Notably, this task lies in a different dimension from other objectives that we expect LLMs to achieve (\textit{e.g.,} improving QA task accuracy) since it is not explicitly included in the pretraining corpora, making it more challenging. Furthermore, the diverse sources and complex distributions of LLM pretraining data, along with varying training corpora for different LLMs, exacerbate the difficulty of this task. Related research is still in its early stages, with existing methods including calibration-based uncertainty estimation~\cite{kadavath2022language} and posterior approaches based on QA testing results~\cite{Yang2023AlignmentFH}, but none of them can satisfactorily solve this issue. For instances near the knowledge boundary, the situation is no longer a binary opposition between knowing and not knowing. In these cases, accurately expressing the model's uncertainty level with fine granularity becomes a critical research question that warrants further exploration.
        

\subsubsection{Minimizing the Alignment Tax During Hallucination Mitigation}
    Several studies have confirmed that the performance of ChatGPT and GPT-4 degrades over time due to continuous safety fine-tuning. This phenomenon is a typical example of an alignment tax. Similarly, when addressing hallucination problems in low-resource languages, the rate of hallucination in the original languages may increase. In the process of enhancing the faithfulness of LLMs' multi-modal perception through visual instruction tuning on VQA datasets, the ``politeness'' of LLM responses may decrease due to the overly succinct nature of VQA answer annotations~\cite{DBLP:journals/corr/abs-2307-01003}. Furthermore, the alignment tax problem is not confined to fine-tuning-based methods but also extends to prompting-based methods such as Retrieval-Augmented Generation (RAG). These methods may also compromise the quality of responses when the quality of retrieved evidence is sub-optimal. In light of these findings, it is clear that we must explore strategies to minimize the cost during hallucination mitigation and prevent LLMs from becoming overly conservative, losing their creativity, or suffering from catastrophic forgetting in task performance.

\subsubsection{Understanding Hallucination in LLMs}
    Although the research community has put extensive efforts into developing empirical methods for measuring and mitigating hallucinations, our understanding of LLM hallucination still remains limited. The absence of reliable theoretical frameworks and rigorous mathematical formulations hinders our ability to answer fundamental questions and make further advancements. For instance, how does hallucination correlate with model and data size (\textit{e.g.}, scaling laws)? What is the relationship between hallucination and auto-regressive generation? Do SFT and RLHF employ different mechanisms to mitigate hallucinations? What is the lower bound of hallucination across different tasks and domains? These questions are also intrinsically linked to a broader question -- although now there are a lot of effort on empirically investigating what LLM can and cannot do from different dimensions, we still don't understand why and how it works. There has been some attempts from various perspectives, such as knowledge encoding~\cite{DBLP:conf/emnlp/GevaSBL21}, compression~\cite{DBLP:journals/corr/abs-2309-10668}, model selection~\cite{DBLP:journals/corr/abs-2311-00871}, compositionality~\cite{DBLP:journals/corr/abs-2305-18654}, computational perspective~\cite{DBLP:journals/corr/abs-2305-15408}, but we are still a long way from truly understanding the mechanism behind various LLM behaviors.

\section{Conclusion}
\label{section:conclusion}
    In this survey, we provide the first comprehensive overview of the hallucination problem in NLG, summarizing existing evaluation metrics, mitigation methods, and the remaining challenges for future research. Hallucination is an artifact of neural-based NLG and is of concern because they appear fluent and can therefore be misleading to users. In some scenarios and tasks, hallucination can cause harm. We survey various contributors to hallucination, ranging from noisy data, erroneous parametric knowledge, incorrect attention mechanism, inappropriate training strategy, to inference exposure bias, etc. We show that there are two categories of hallucinations, namely intrinsic hallucination and extrinsic hallucination, and they need to be treated differently with diverse mitigation strategies. Hallucination is relatively easy to detect in abstractive summarization and in NMT against the evidence in the source. For dialogue systems, it is important to balance diversity vs consistency in dialogue responses. Hallucination in GQA and VL tasks is detrimental to the performance, but research on mitigation methods is still very preliminary in these areas. For data-to-text generation, hallucination arises from the discrepancy between the input and output format. Most methods to mitigate hallucinations in NMT either aim to reduce dataset noise or alleviate exposure bias. In the VL domain, models also generate unfaithful output given the visual scene, and recent works have mainly focused on the object hallucination problem. There remain many challenges ahead in identifying and mitigating hallucinations in NLG, and we hope research in this area can benefit from this survey.

\bibliographystyle{ACM-Reference-Format}
\bibliography{sample-acmsmall}


\end{document}